\documentclass[10pt,twocolumn,letterpaper]{article}

\usepackage{cvpr}      

\usepackage[pagebackref,breaklinks,colorlinks,allcolors=cvprblue]{hyperref}

\usepackage{graphicx}      
\usepackage{amsmath}       
\usepackage{amssymb}       
\usepackage{booktabs}      
\usepackage{multirow}      
\usepackage{subcaption}    

\usepackage{diagbox}       
\usepackage{array}         
\usepackage{adjustbox}     
\usepackage{makecell}      

\usepackage{algorithm}
\usepackage{algorithmic}

\usepackage{newfloat}
\usepackage{listings}
\DeclareCaptionStyle{ruled}{labelfont=normalfont,labelsep=colon,strut=off}
\floatstyle{ruled}
\newfloat{listing}{tb}{lst}{}
\floatname{listing}{Listing}

\usepackage[table,x11names]{xcolor}
\definecolor{first}{HTML}{F48288}
\definecolor{second}{HTML}{FAC791}
\definecolor{third}{HTML}{FFFF8F}
\definecolor{gray}{HTML}{D3D3D3}
\definecolor{urlblue}{RGB}{0,112,245} 

\usepackage[capitalize,noabbrev]{cleveref}

\definecolor{cvprblue}{rgb}{0.21,0.49,0.74}

\def\confName{CVPR}
\def\confYear{2026}

\title{MAPo : Motion-Aware Partitioning of Deformable 3D Gaussian Splatting for High-Fidelity Dynamic Scene Reconstruction}

\author{
Han Jiao, Jiakai Sun\\
Zhejiang University\\
{\tt\small \{csjh, csjk\}@zju.edu.cn}
\and
Yexing Xu\\
The Shenzhen Campus, Sun Yat-Sen University\\
{\tt\small xuyx55@mail2.sysu.edu.cn}
\and
Lei Zhao, Wei Xing, Huaizhong Lin\\
Zhejiang University\\
{\tt\small \{cszhl, wxing, linhz\}@zju.edu.cn}
}

\begin{document}
\twocolumn[{
\renewcommand\twocolumn[1][]{#1}
\maketitle
\begin{center}
    \captionsetup{type=figure}
    \includegraphics[width=\textwidth,
      trim={0 0 0 0},  
    clip]{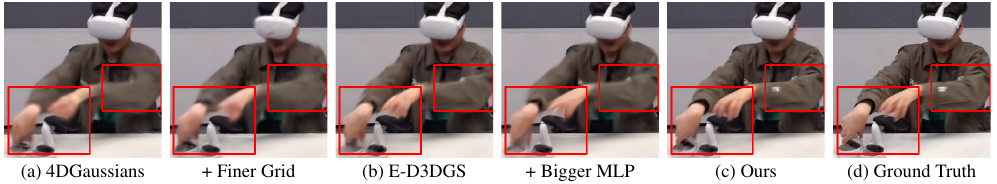}
    
    \vspace{-0.5mm}
    \captionof{figure}{\textbf{Overview.} (a-b) Deformation-based methods often blur details in regions with complex or rapid motion. (c) Our MAPo significantly improves rendering quality in these areas. (d) Ground Truth.}
    \label{fig:teaser_overview}
\end{center}
}]
\begin{abstract}
3D Gaussian Splatting, known for enabling high-quality static scene reconstruction with fast rendering, is increasingly being applied to multi-view dynamic scene reconstruction. A common strategy involves learning a deformation field to model the temporal changes of a canonical set of 3D Gaussians. However, these deformation-based methods often produce blurred renderings and lose fine motion details in highly dynamic regions due to the inherent limitations of a single, unified model in representing diverse motion patterns. To address these challenges, we introduce Motion-Aware Partitioning of Deformable 3D Gaussian Splatting (MAPo), a novel framework for high-fidelity dynamic scene reconstruction. Its core is a dynamic score-based partitioning strategy that distinguishes between high- and low-dynamic 3D Gaussians. For high-dynamic 3D Gaussians, we recursively partition them temporally and duplicate their deformation networks for each new temporal segment, enabling specialized modeling to capture intricate motion details. Concurrently, low-dynamic 3DGs are treated as static to reduce computational costs. However, this temporal partitioning strategy for high-dynamic 3DGs can introduce visual discontinuities across frames at the partition boundaries. To address this, we introduce a cross-frame consistency loss, which not only ensures visual continuity but also further enhances rendering quality. Extensive experiments demonstrate that MAPo achieves superior rendering quality compared to baselines while maintaining comparable computational costs, particularly in regions with complex or rapid motions. 
\end{abstract}    
\section{Introduction}
\label{sec:intro} 
Reconstructing high-fidelity dynamic scenes from multi-view video inputs is a fundamental challenge in computer vision, with broad applications in virtual reality, visual effects, and autonomous driving. In recent years, Neural Radiance Fields (NeRF)~\cite{NeRF}  have demonstrated remarkable capabilities in representing static scenes for novel view synthesis by leveraging implicit neural representations. Building upon this, numerous efforts~\cite{D-NeRF, SpacetimeNerf, Hypernerf, Nerfies, StreamRF, DyNeRF, TiNeuVox, K-Planes, Hexplane, tensor4d} have extended NeRF to dynamic scenes by introducing temporal conditioning. However, the inherent reliance on dense spatial sampling and costly Multilayer Perceptron (MLP) querying leads to significant limitations in both training efficiency and rendering speed.
\par
Recently, 3D Gaussian Splatting (3DGS)~\cite{3DGS} has emerged as a powerful alternative for static scene reconstruction, achieving real-time rendering while maintaining photorealistic rendering quality. This method explicitly represents scenes using anisotropic 3D Gaussian (3DG) primitives. Encouraged by its success in static scenes, researchers have begun adapting Gaussian Splatting to dynamic scenes~\cite{deformable3dgs, 4dgs_kplanes, SC-GS, ed3dgs, wu2025swift4dadaptivedivideandconquergaussiansplatting, dn4dgs, ex4dgs, realtime4dgs,wu2025localdygsmultiviewglobaldynamic}. Employing deformable 3D Gaussians—where a learned deformation field maps a canonical set of 3D Gaussians (3DGs) to their time-varying states—has quickly emerged as a prevailing and effective strategy for reconstructing dynamic scenes, offering compelling visual quality from a compact representation. Despite promising results, these methods suffer from two critical limitations inherent in their deformation framework:
\begin{figure}[t]
    \centering

    \begin{subfigure}{0.32\linewidth}
        \centering
        \includegraphics[
            width=\linewidth,
        ]{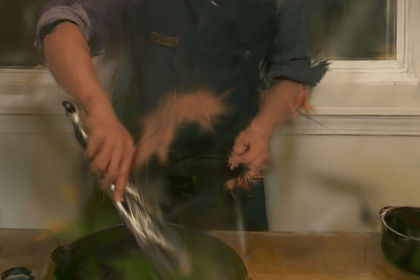}
        \caption*{(a) Average}
    \end{subfigure}
    \hfill
    \begin{subfigure}{0.32\linewidth}
        \centering
        \includegraphics[
            width=\linewidth,
        ]{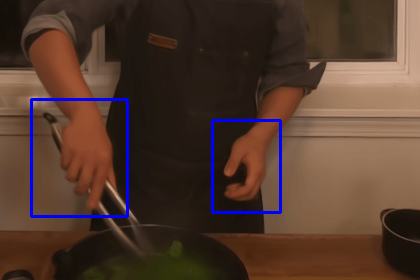}
        \caption*{(b) Rendering 1}
    \end{subfigure}
    \hfill
    \begin{subfigure}{0.32\linewidth}
        \centering
        \includegraphics[
            width=\linewidth,
        ]{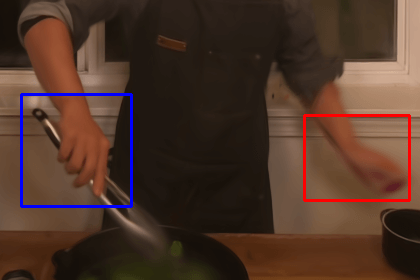}
        \caption*{(c) Rendering 2}
    \end{subfigure}
    
    \caption{\textbf{Rendering results of a single unified model.} 
    (a) shows the temporally averaged representation, which is visualized by directly rendering the canonical 3DGs. The regions highlighted in \textcolor{blue}{blue} in (b) and (c) are visually close to this average. The region highlighted in \textcolor{red}{red} in (c) is visually distant from this average.
    }
    \label{fig:intro_avg_figs}
    \vspace{-1mm}
\end{figure}
\begin{itemize}
    \item \textbf{Bottleneck in Motion Modeling Capacity:} As shown in Fig.~\ref{fig:teaser_overview}, deformation-based methods often produce blurry reconstructions and lose fine motion details in regions with complex or rapid motions. 
    This core limitation stems from their unified modeling strategy, which relies on a single canonical set of 3DGs and a single, globally shared deformation network to represent all spatio-temporal variations. Such an approach forces the network to find a single set of parameters that best fits all diverse and often conflicting motion patterns across the entire time sequence. Consequently, the learned representation often converges to a temporally averaged representation. As shown in  Fig.~\ref{fig:intro_avg_figs}, this averaging effect prevents the model from accurately capturing complex motions that deviate from this learned average, such as abrupt changes and fine-grained temporal details. This ultimately leads to the loss of fine details in dynamic regions.
    \item \textbf{Redundant Computation:} The 3DGs in static regions still participate in repeated deformation network computation, wasting computational resources and potentially slowing down training and rendering speed.
\end{itemize}
To tackle these issues, we introduce MAPo, a novel framework for high-fidelity dynamic scene reconstruction. Its core is a dynamic score-based partitioning strategy comprising two key components: Temporal Partitioning Based on Dynamic Scores and Static 3D Gaussian Partitioning. The former enhances details in dynamic regions through hierarchical temporal partitioning of high-dynamic 3DGs, and the latter resolves redundant computation by treating low-dynamic 3DGs as static 3DGs. To ensure temporal smoothness after partitioning at the partition boundaries, a cross-frame consistency loss is also employed.
\par
Specifically, MAPo first analyzes each 3DG's historical positions to compute a dynamic score, which provides a practical measure of motion intensity for each 3DG. During temporal partitioning, high-dynamic 3DGs, identified by their high dynamic scores, are recursively partitioned along the temporal dimension. The 3DGs that undergo partitioning are replicated, creating new 3DG instances for the resulting temporal sub-segments. Simultaneously, the deformation network of the parent segment is replicated, creating a dedicated sub-network for each new temporal sub-segment to deform all 3DGs within it. Instead of relying on a single, unified model, our strategy adaptively enables multiple sets of networks and their corresponding 3DGs to model the dynamic scene over distinct temporal segments. This allows for a more faithful capture of details in highly dynamic regions and effectively alleviates the ``temporal averaging" effect.
Meanwhile, 3DGs with low dynamic scores are identified as static. Their attributes are directly updated to reflect the deformation results, after which redundant deformation network computations are skipped to reduce computational costs. 

To further mitigate visual discontinuities introduced by temporal partitioning, we introduce a cross-frame consistency loss that enforces two constraints: (i) the renderings of two sets of temporally adjacent 3DGs should remain as consistent as possible when rendered at the same time and from the same viewpoint, and (ii) the rendering of the temporally adjacent 3DGs corresponding to the current time should also align with the ground-truth observation at the current time. This consistency loss not only reduces visual discontinuities at partition boundaries but also leverages temporal context for more faithful reconstructions. Our key contributions are summarized as follows:

\begin{itemize}
    \item We propose MAPo, a novel framework for high-fidelity dynamic scene reconstruction based on a dynamic score-based partitioning strategy. Our strategy enhances the modeling effects of highly dynamic regions by enabling multiple sets of networks and their corresponding 3DGs to model the dynamic scene over distinct temporal segments, and simultaneously reduces computational costs by identifying low dynamic scores as static.
    \item We design a cross-frame consistency loss that effectively mitigates visual discontinuities introduced by temporal partitioning and further enhances the rendering quality.
    \item Extensive experiments demonstrate that MAPo achieves state-of-the-art (SOTA) rendering quality, particularly in capturing fine details in highly dynamic scenes.
\end{itemize}
\section{Related Work}
\label{sec:related_work}
\subsection{Dynamic NeRF}


Neural Radiance Fields (NeRF)~\cite{NeRF} marked a significant breakthrough in representing static scenes for high-quality novel view synthesis using implicit MLPs. This success has spurred numerous efforts to extend its capabilities to dynamic scenes.
One prominent strategy is deformation-based, where a learned deformation field maps observed points to a canonical representation. D-NeRF~\cite{D-NeRF} pioneered this concept by directly using spatio-temporal coordinates as the input to the deformation network. This was later advanced by methods like Nerfies~\cite{Nerfies}, replacing the explicit time input with a learnable latent code to govern deformation, and HyperNeRF~\cite{Hypernerf}, utilizing higher-dimensional embeddings to better model complex topologies. NeRFPlayer~\cite{nerfplayer} introduces a streaming-capable representation that separates static and dynamic scene components.
Other approaches bypass the deformation-based paradigm. For instance, DyNeRF~\cite{DyNeRF} directly queries the appearance attributes of spatio-temporal points using a 6D spatio-temporal function. 
To improve efficiency, K-Planes~\cite{K-Planes} and HexPlane~\cite{Hexplane} introduce explicit multi-planar representations. HyperReel~\cite{hyperreel} further enhances rendering speed by employing a more compact dual-plane representation.
Despite these advances, the reliance on dense ray sampling and per-point MLP queries in NeRF-based methods remains a barrier to achieving real-time rendering.

\subsection{Dynamic Gaussian Splatting}
The recent success of 3DGS in achieving real-time, photorealistic rendering for static scenes has naturally inspired its extension to dynamic scenes.
Among these extensions, a significant number of approaches are deformation-based, where a canonical set of 3DGs is deformed over time. D3DGS~\cite{deformable3dgs} introduces a canonical 3DGs representation and a deformation network that takes position and time as input to map canonical 3DGs to the observation space. Many subsequent methods share similarities with this paradigm. 4DGaussians~\cite{4dgs_kplanes} decodes features from a HexPlane representation using spatio-temporal coordinates to derive per-Gaussian deformations. E-D3DGS~\cite{ed3dgs} advances this by replacing direct coordinate inputs with temporal and per-Gaussian embeddings and adopting a dual-deformation strategy. Similarly, DN-4DGS~\cite{dn4dgs} integrates spatio-temporal information from neighboring elements and also employs a dual-deformation approach. To handle long sequences, SWinGS~\cite{shaw2024swingsslidingwindowsdynamic} partitions the sequence into sliding windows and trains an independent model for each, using 2D optical flow to guide window segmentation and a post-hoc fine-tuning step to ensure smoothness. To improve efficiency, Swift4D~\cite{wu2025swift4dadaptivedivideandconquergaussiansplatting} uses 2D RGB information to separate dynamic from static regions, modeling only the dynamic 3DGs' deformation with a 4D hash grid and a small MLP decoder. Inspired by Scaffold-GS~\cite{scaffoldgs}, LocalDyGS~\cite{wu2025localdygsmultiviewglobaldynamic} decomposes the scene into seed-based local spaces and generates time-varying 3DGs by fusing static and dynamic features, though its overall quality is still limited. While diverse, these deformation-based methods face a fundamental trade-off. Approaches relying on a single, globally shared representation like D3DGS and its direct successors often suffer from the ``temporal averaging" effect in complex scenes. Conversely, strategies like SWinGS that partition the sequence into independent, localized models mitigate this issue, but introduce their own challenges, including coarse, window-level partitioning; cumbersome pre- and post-processing steps; and a strong reliance on 2D priors like optical flow.

Other distinct strategies have also been explored. Curve-based methods~\cite{gaussianflow, SpacetimeGaussians, ex4dgs} model 3DG attribute changes over time using parametric curves, but may induce drift under complex motions and introduce significant storage overhead. 4D Gaussian-based methods ~\cite{4dgs_towards, realtime4dgs, Xu_2024} decompose 4D Gaussians into 3DGs and marginal 1D Gaussians, offering a direct spatio-temporal representation, but incur significant computational costs. Per-frame training methods~\cite{3dgstream, hicom,4DGC} enable online reconstruction but are constrained by the inherent trade-off between quality, storage, and efficiency.

To overcome the limitations of existing methods, we introduce a dynamic score-based partitioning strategy. We recursively partition high-dynamic 3DGs into finer temporal segments to address the issue of the ``temporal average". Unlike window-based methods like SWinGS, our partitioning operates at a fine-grained, per-3DG level, is guided directly by 3D motion instead of 2D priors, and is integrated into a single, end-to-end training framework, thus avoiding cumbersome pre- and post-processing steps. Furthermore, we build our method upon a compact deformation field architecture and identify low-dynamic 3DGs as static to avoid the substantial overhead of other categories of methods.
\begin{figure*}
    \centering
    \includegraphics[width=\linewidth]{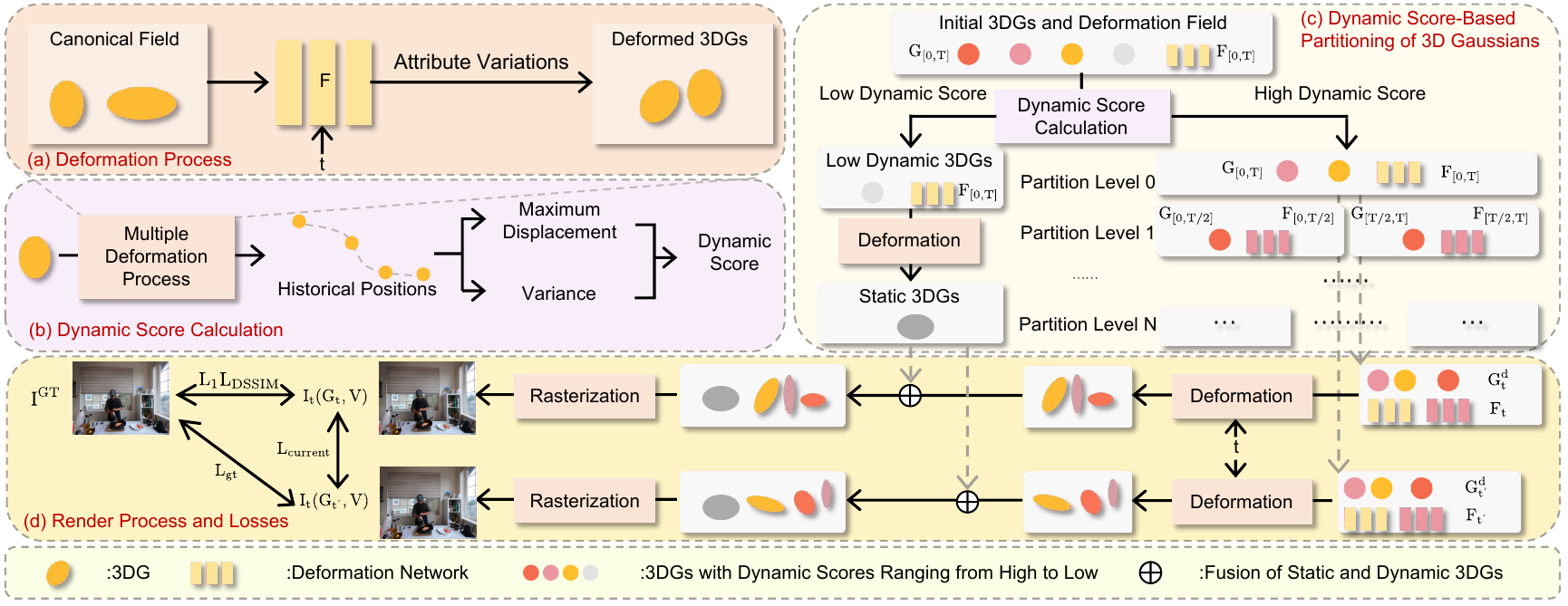}
    \vspace{-2mm}
    \caption{\textbf{An overview of MAPo.} (a) 3DGs' deformation process.
    (b) Compute the dynamic score of 3DGs from history positions during training.
    (c) High-dynamic 3DGs are recursively temporally partitioned, and low-dynamic ones are deformed and treated as static.
    (d) Dynamic and static 3DGs are combined for rendering. Losses are computed on the left.
        }
        \vspace{-3mm}
\label{fig: framework}
\end{figure*}
\section{Preliminaries}
\label{sec:preliminaries}
\paragraph{3D Gaussian Splatting}
3D Gaussian Splatting introduces an explicit point-based representation where each point in the point cloud is equipped with four fundamental properties: mean $\mu$, covariance matrix $\Sigma$, opacity $\alpha$, and spherical harmonics $sh$. The covariance matrix $\Sigma$ can be decomposed as $\Sigma = RSS^{T}R^{T}$, where $S$ is parameterized by a 3D vector $s$ and $R$ is parameterized by a quaternion $q$. The rendering process is fully differentiable. For a given viewpoint, all 3DGs are first projected onto the 2D image plane. Then, these projected 2D Gaussians are sorted by depth and blended together using alpha-compositing to synthesize the final pixel color $C$:
\begin{equation}
    C = \sum_{i=1}^{N} c_{i} \alpha_{i}' \prod_{j=1}^{i-1} \left(1 - \alpha_{j}'\right).
\label{eq:alpha_compositing}
\end{equation}
where $c_i$ is the view-dependent color computed from $sh$ and viewpoint, and $\alpha'_i$ is the 2D evaluation of the 3DG's opacity.


\paragraph{Embedding-Based Deformation}
Our work builds upon the dual deformation paradigm introduced in E-D3DGS~\cite{ed3dgs}. In this paradigm, each 3DG is associated with a learnable embedding $\mathbf{z}_g$, and each timestamp $t$ is represented by a pair of temporal embeddings: a coarse embedding $\mathbf{z}_{t_c}$ capturing low-frequency motion, and a fine embedding $\mathbf{z}_{t_f}$ for high-frequency details.
A coarse deformation network $\mathcal{F}$ and a fine deformation network $\mathcal{F}_{\theta}$ process these embeddings to predict deformations for all 3DG attributes. The final deformation is the sum of the coarse and fine predictions:
\begin{equation}
    (\Delta {\mu}, \Delta {q}, \Delta {s}, \Delta {\alpha}, \Delta {sh}) = \mathcal{F}(\mathbf{z}_g, \mathbf{z}_{t_c}) + \mathcal{F}_{\theta}(\mathbf{z}_g, \mathbf{z}_{t_f}).
\label{eq:coarse_fine_deformation}
\end{equation}

\section{Method}
\label{sec:method}

Our approach consists of two main components: a dynamic score-based partitioning strategy and a cross-frame consistency loss. The overview of our method is shown in Fig.~\ref{fig: framework}. 
First, we elaborate on the dynamic score-based partitioning strategy, including how we compute a dynamic score for each 3DG based on its historical positions, and how this score guides a partitioning where high-dynamic 3DGs are recursively temporally partitioned while low-dynamic 3DGs are identified as static.
Subsequently, we describe our cross-frame consistency loss, which is designed to address the visual discontinuities caused by partitioning.

\subsection{Dynamic Score-based Partitioning Strategy}
\subsubsection{Dynamic Score Calculation}
To accurately characterize the motion intensity of each 3DG, we design a comprehensive scoring mechanism that integrates both maximum displacement and position variance. For each 3DG \(G_i\), we record its spatial position \(\mu_{ij}\) during training, where \(i\in\{1,2,\cdots,N\}\) denotes the index of 3DG and \(j\in\{1,2,\cdots,m\}\) represents the index of the recorded historical positions. Here, \(m\) is a hyperparameter controlling the number of recorded positions per 3DG. 
Initially, we considered using the maximum displacement as the sole metric for quantifying the motion intensity of 3DGs, which we efficiently compute as the length of the diagonal of the historical positions' axis-aligned bounding box. However, upon closer examination of the positional records, we realized that relying solely on this metric would be insufficient. For instance, an object with short-term high-speed motion but long-term stillness may exhibit a large maximum displacement but relatively low variance. Conversely, an object with continuous small oscillations would have a small maximum displacement but a relatively high variance, suggesting a consistent and complex yet less pronounced movement pattern. Therefore, to comprehensively characterize diverse motion behaviors to accurately identify more challenging motion, we employ two quantitative metrics for each 3DG. For the i-th 3DG, its maximum displacement $r_i$ and position variance $v_i$ are computed as:
\begin{equation}
  r_i = \bigl\|\max_j \mu_{ij} - \min_j \mu_{ij}\bigr\|, \quad
  v_i = \sum_{j=1}^{m} \frac{\|\mu_{ij} - \bar{\mu}_i\|^2}{m},
\end{equation}
Here, $\max_j \mu_{ij}$ and $\min_j \mu_{ij}$ are the element-wise maximum and minimum vectors over the historical positions $\{\mu_{ij}\}_{j=1}^m$ for 3DG $i$. The term $\bar{\mu}_i = \frac{1}{m} \sum_{j=1}^{m} \mu_{ij}$ represents the mean position. The maximum displacement $r_i$ captures the peak amplitude of motion, while the variance $v_i$ measures the dispersion around the mean position. After computing the maximum displacement and variance, we map all 3DGs' $r$ and $v$ values to the interval $[0,1]$ via percentile-based normalization:
\begin{equation}
  \tilde r_i = \sum_{k=1}^{100} \frac{\mathbf{1}(r_i \ge q_r(k))}{100}, \quad
  \tilde v_i = \sum_{k=1}^{100} \frac{\mathbf{1}(v_i \ge q_v(k))}{100}, 
\end{equation}
where $\mathbf{1}(\cdot)$ denotes the indicator function, and $q_r(k)$ and $q_v(k)$ are the $k$-th percentiles of $\{r_i\}$ and $\{v_i\}$, respectively.
We use the harmonic mean to fuse \(\tilde{r}_i\) and \(\tilde{v}_i\), as it requires both inputs to be high for a high output. The final dynamic score $S_i$ of the i-th 3DG is denoted as:
\begin{equation} S_i = \frac{2}{\frac{1}{\tilde r_i + \varepsilon} + \frac{1}{\tilde v_i + \varepsilon}},\end{equation}
where $\varepsilon = 10^{-6}$ is used for numerical stability. 
\subsubsection{Temporal Partitioning Based on Dynamic Scores} 
\begin{figure}
    \centering
    \begin{subfigure}{0.23\linewidth}
        \centering
        \includegraphics[width=\linewidth, trim=30 10 5 0, clip]{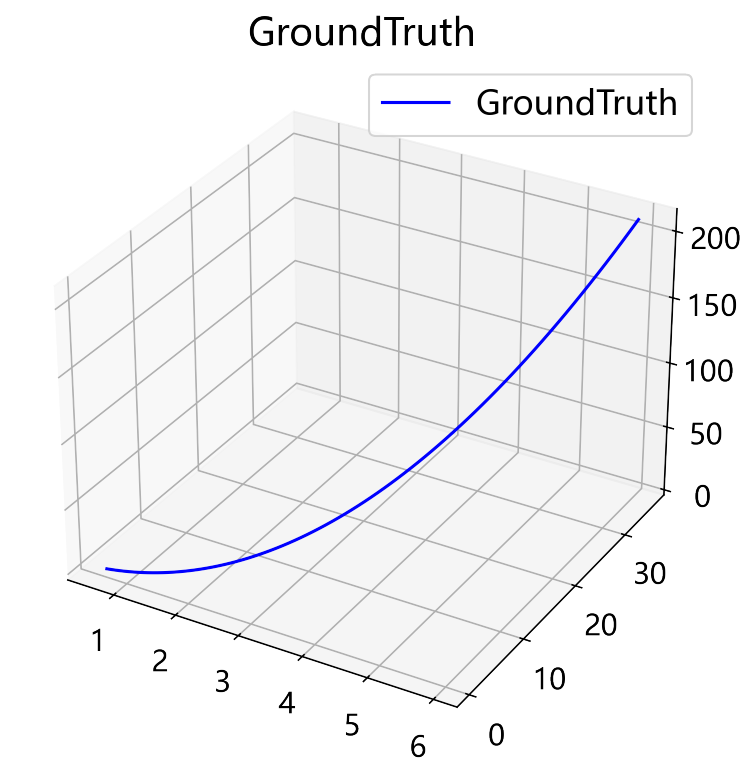}
        \caption*{(a)}
    \end{subfigure}
    \hfill
    \begin{subfigure}{0.24\linewidth}
        \centering
        \includegraphics[width=\linewidth, trim=30 10 20 10, clip]{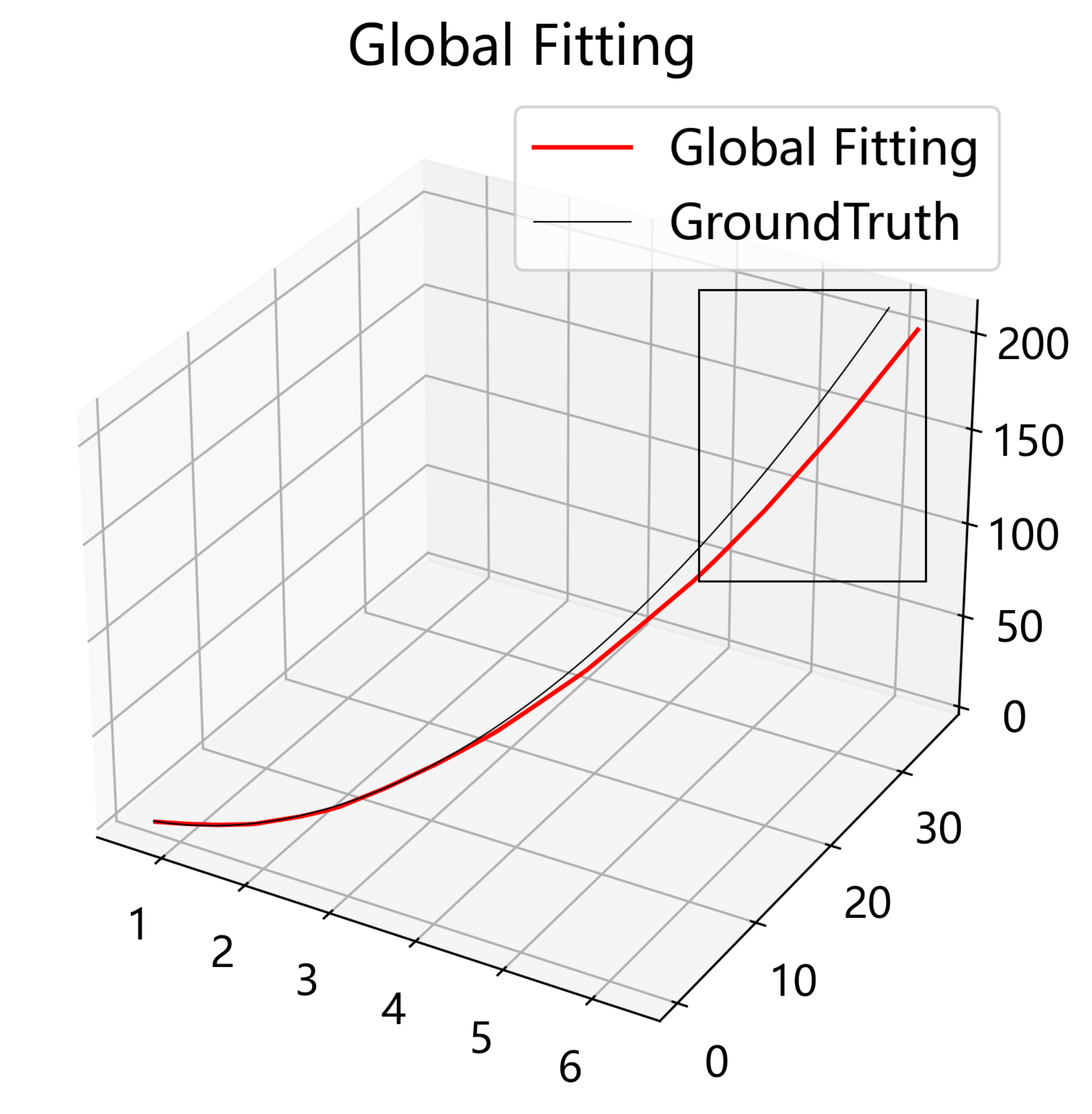}
        \caption*{(b)}
    \end{subfigure}
    \hfill
    \begin{subfigure}{0.24\linewidth}
        \centering
        \includegraphics[width=\linewidth, trim=30 10 20 10, clip]{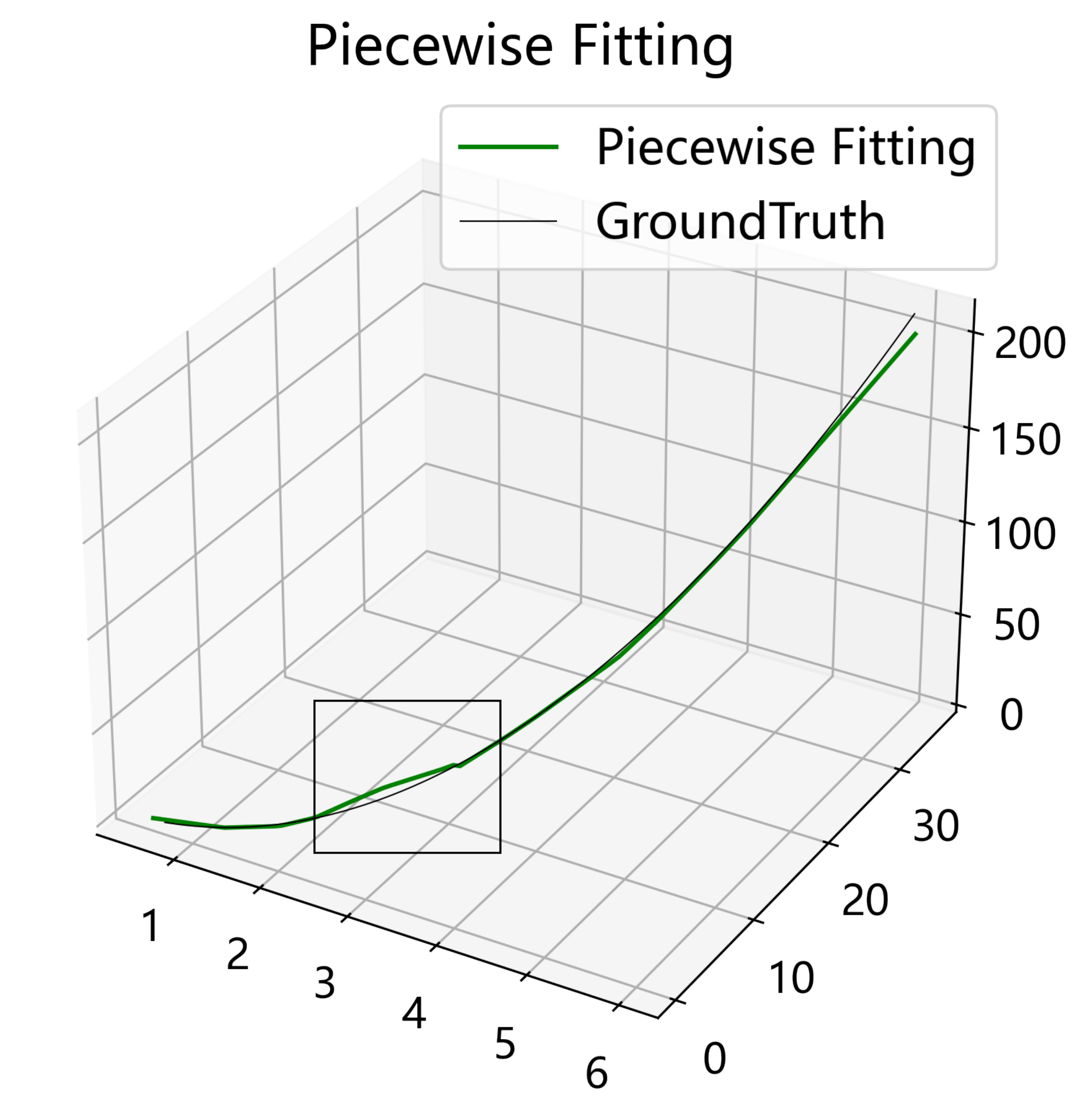}
        \caption*{(c)}
    \end{subfigure}
    \hfill
    \begin{subfigure}{0.23\linewidth}
        \centering
        \includegraphics[width=\linewidth, trim=30 10 20 10, clip]{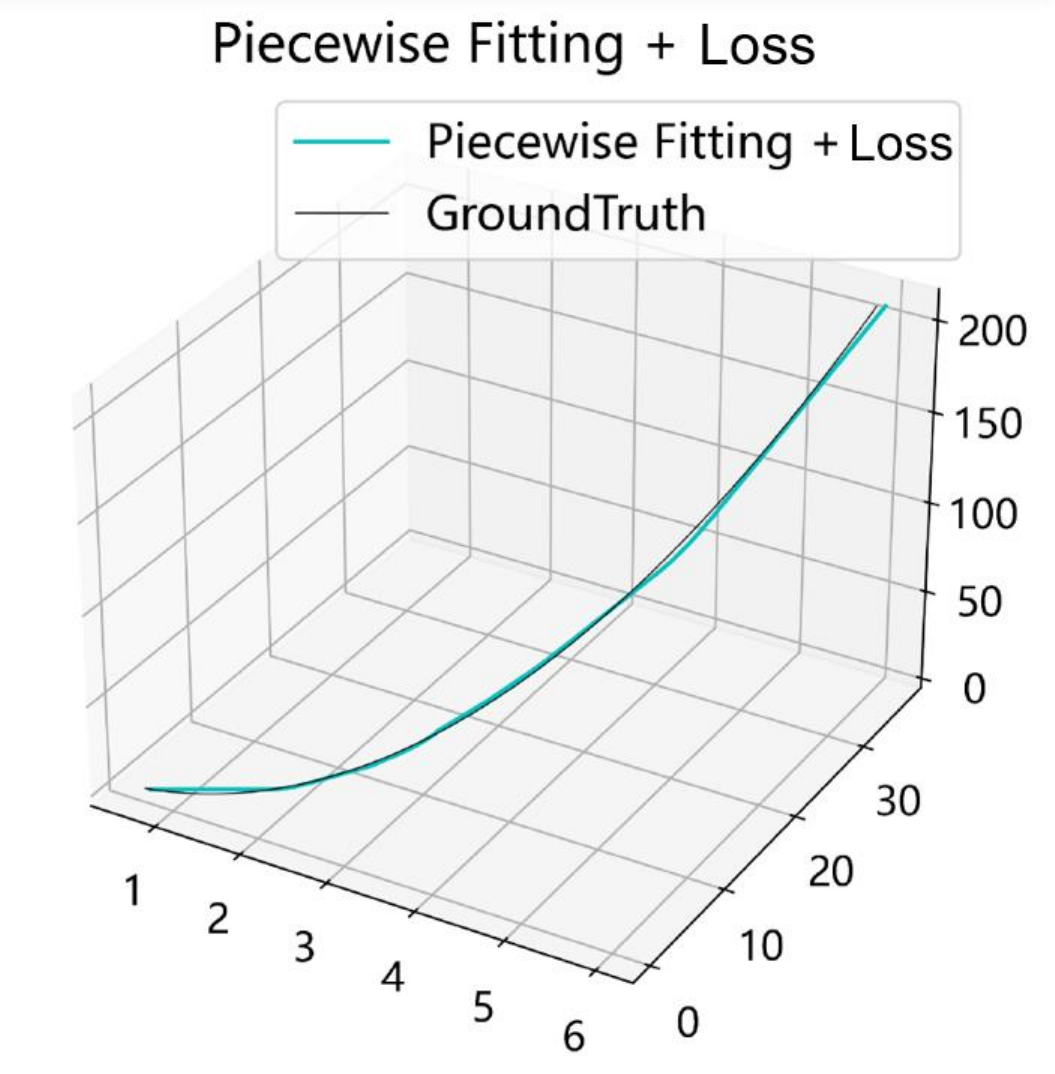}
        \caption*{(d)}
    \end{subfigure}
    \vspace{-2mm}
    \caption{\textbf{Effectiveness of temporal partitioning strategy and consistency loss on a toy example.}
(a) A 3D curve \(\mathbf{p}(t)\) simulates a dynamic trajectory.
(b) A single point and a single MLP to fit \(\mathbf{p}(t)\) for the entire duration;
(c) Two points and two corresponding MLPs for two partitioned time segments;
(d) Apply a consistency loss to (c) at the partition boundary.
}
    \label{fig:toy_demo}
    \vspace{-3mm}
\end{figure}

3DGs with high dynamic scores typically correspond to regions with complex or rapid motion. Since a single 3DG struggles to effectively model long-term or complex motion characteristics, an intuitive and effective solution is to partition these high-dynamic 3DGs along the temporal dimension based on their dynamic scores, thereby capturing motion variations more precisely. Fig.~\ref{fig:toy_demo} demonstrates the effectiveness of this partitioning through a simple demo.

\par
Building on this insight, we recursively partition the 3DGs along the temporal dimension, progressively subdividing those with higher dynamic scores into finer temporal segments to achieve more precise dynamic reconstruction. 
Specifically, for the complete time range $[0,T]$, each 3DG maintains two key properties: its partition level $l$ (initialized to 0) and its temporal segment range $[t_\text{start}, t_\text{end}]$ (initially $[0,T]$). For notational convenience, all such ranges are treated as left-closed and right-open. Let $G_{[t_\text{start},t_\text{end}]}$ denote the set of 3DGs assigned to the temporal segment range $[t_\text{start}, t_\text{end}]$. Let $F_{[t_\text{start},t_\text{end}]}$ represent the deformation network corresponding to the temporal segment range $[t_\text{start}, t_\text{end}]$. It is responsible for deforming the 3DGs in $G_{[t_\text{start},t_\text{end}]}$. 
When a 3DG at level $l$ exhibits a dynamic score exceeding the current-level threshold $\tau_l$ within its time range $[t_\text{start}, t_\text{end}]$, we partition it at the temporal midpoint $t_\text{mid} = (t_\text{start} + t_\text{end})/2$. The original 3DG retains the first sub-segment $[t_\text{start}, t_\text{mid}]$ while advancing to level $l + 1$, and a new replica is created for the second sub-segment $[t_\text{mid}, t_\text{end}]$ with identical attributes. 
Correspondingly, the deformation network $F_{[t_\text{start},t_\text{end}]}$ is replicated to create $F_{[t_\text{start},t_\text{mid}]}$ and $F_{[t_\text{mid},t_\text{end}]}$ to model the distinct spatio-temporal deformation patterns within each sub-segment. Within each new sub-segment, this partitioning process is applied recursively. 

\subsubsection{Static 3D Gaussian Partitioning}
3DGs with dynamic scores below a predefined threshold $\tau_{\text{static}}$ are identified as static. These static 3DGs have their attributes initialized once using the output of their associated deformation network at a randomly sampled timestep. Subsequently, they are excluded from computations involving the deformation network during rendering while their attributes remain optimizable, significantly reducing computational costs. 

\begin{figure*}[!htbp]
    \centering

    \begin{subfigure}{0.16\textwidth}
      \includegraphics[width=\linewidth]{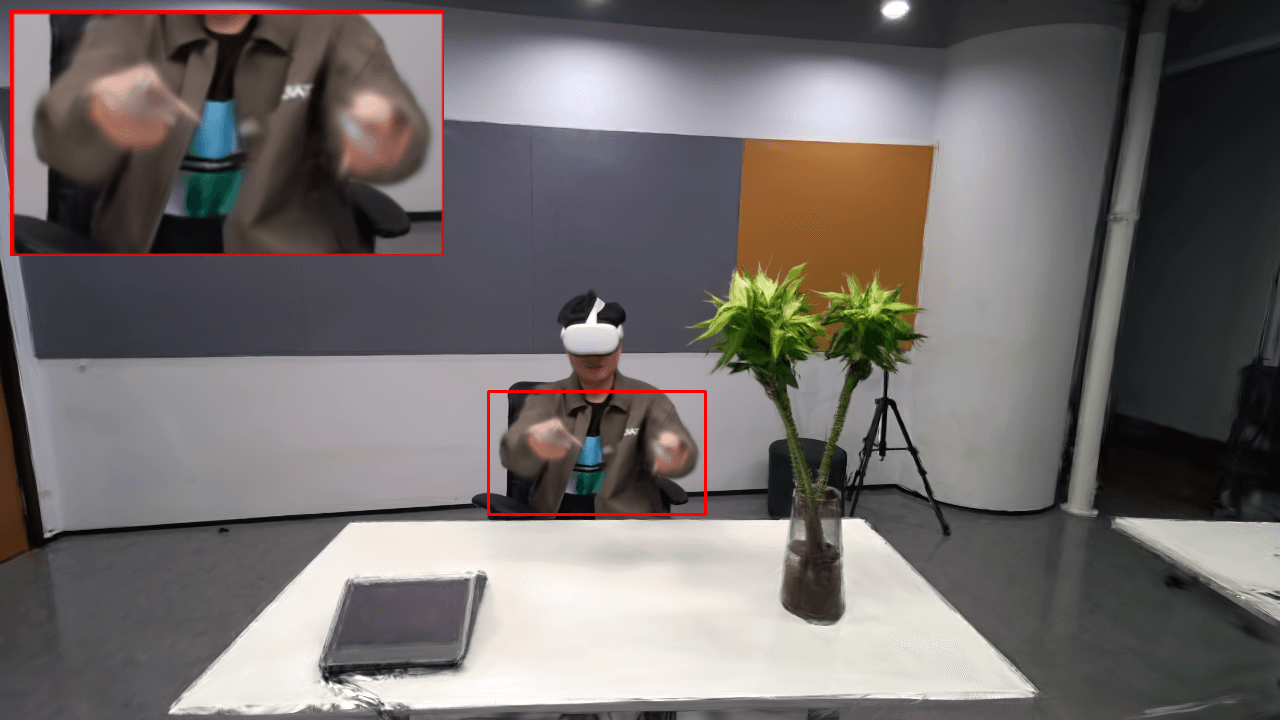}
    \end{subfigure}
    \begin{subfigure}{0.16\textwidth}
        \includegraphics[width=\linewidth]{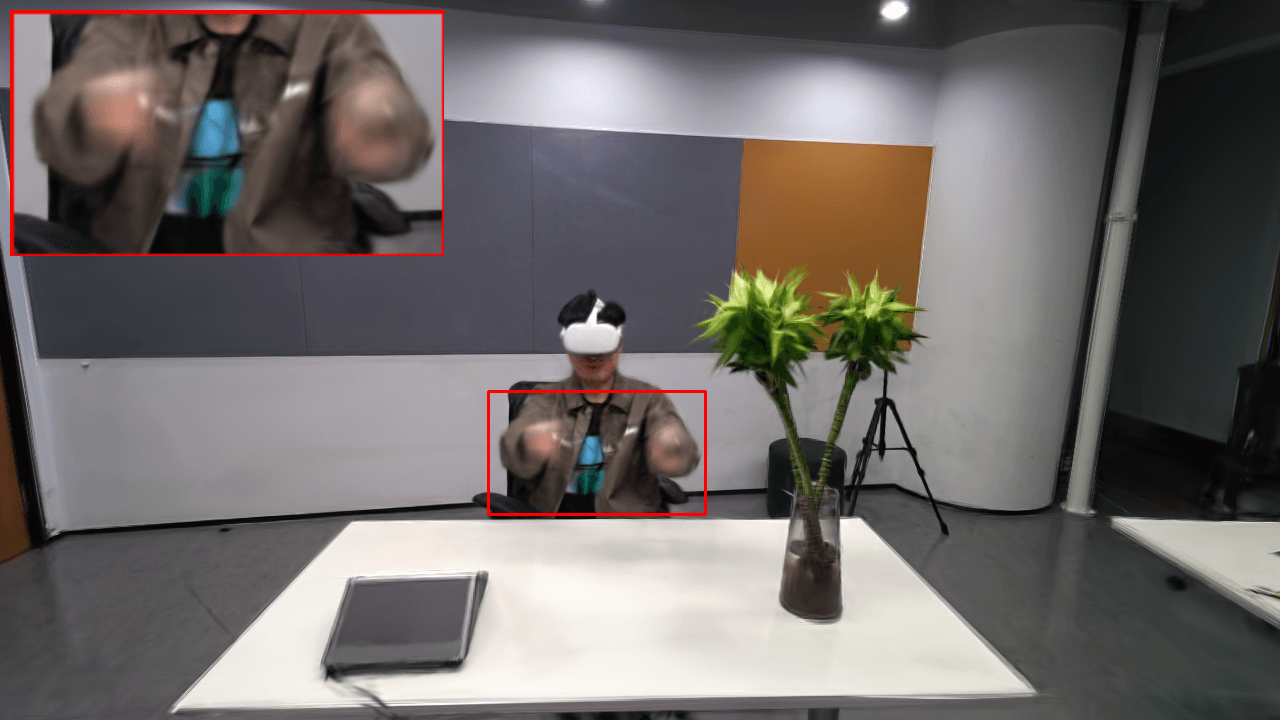}
    \end{subfigure}
    \begin{subfigure}{0.16\textwidth}
        \includegraphics[width=\linewidth]{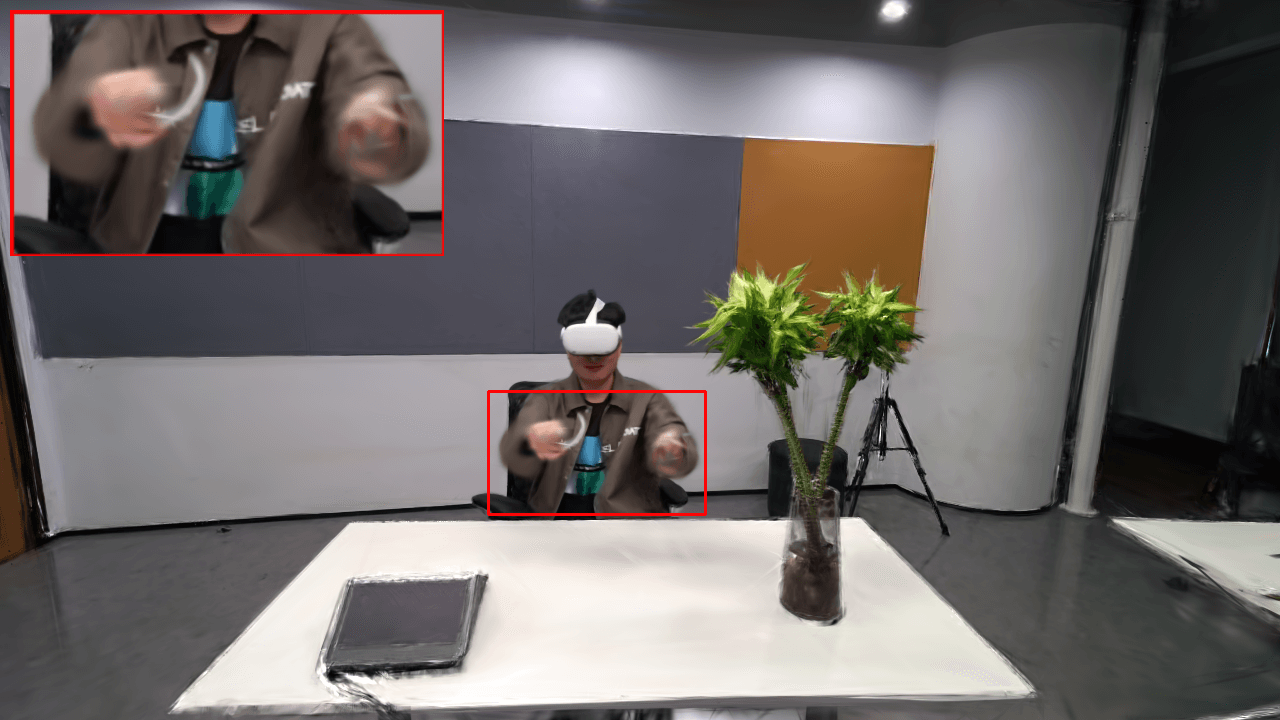}
    \end{subfigure}
    \begin{subfigure}{0.16\textwidth}
        \includegraphics[width=\linewidth]{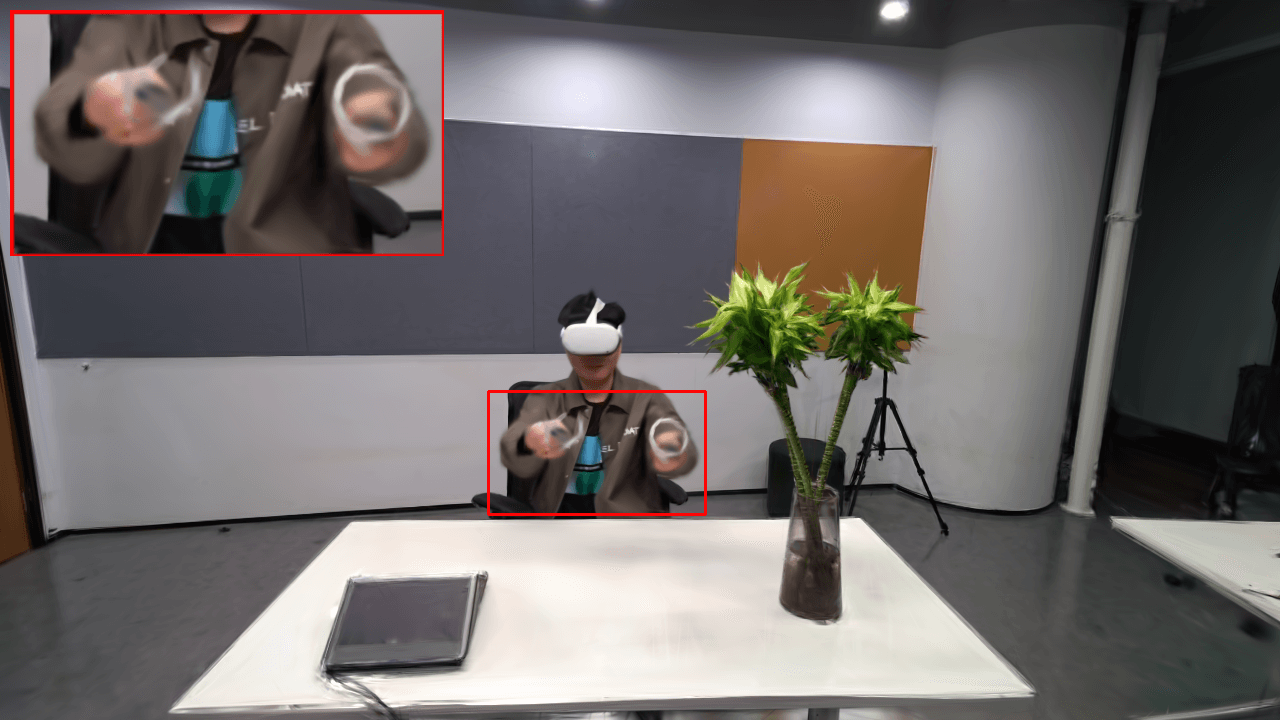}
    \end{subfigure}
    \begin{subfigure}{0.16\textwidth}
        \includegraphics[width=\linewidth]{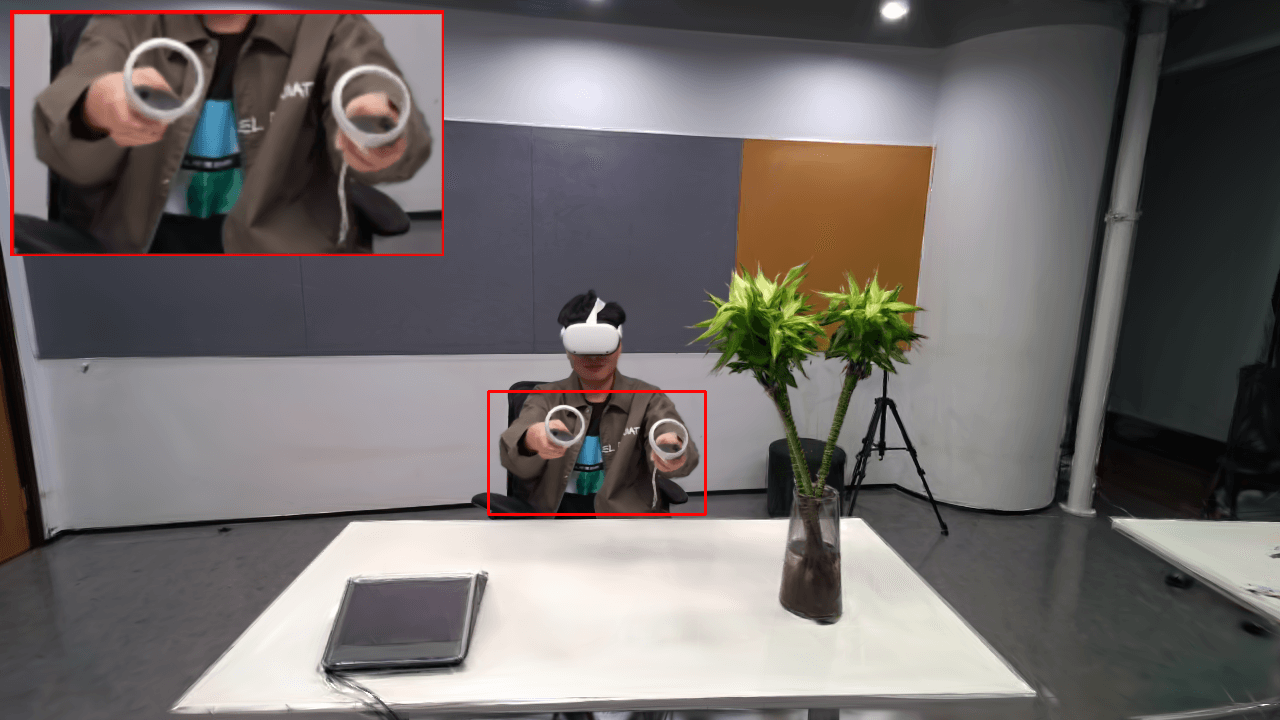}
    \end{subfigure}
    \begin{subfigure}{0.16\textwidth}
        \includegraphics[width=\linewidth]{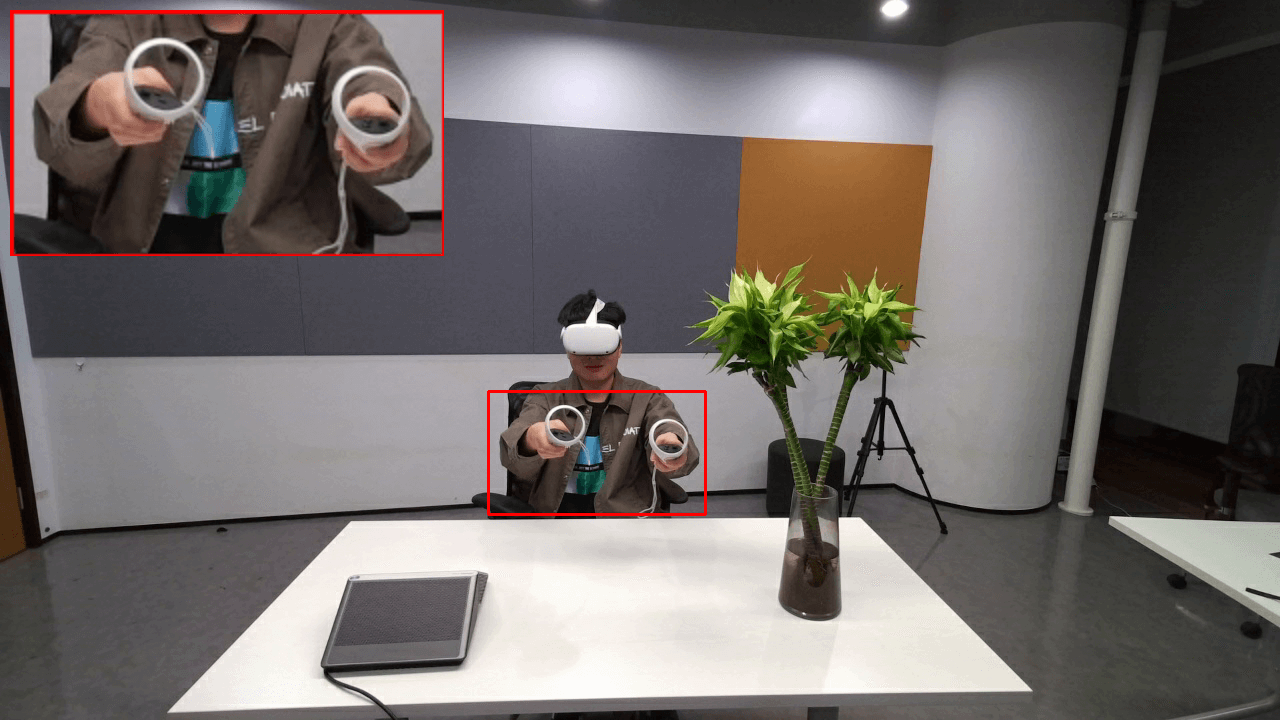}
    \end{subfigure}

    \begin{subfigure}{0.16\textwidth}
      \includegraphics[width=\linewidth]{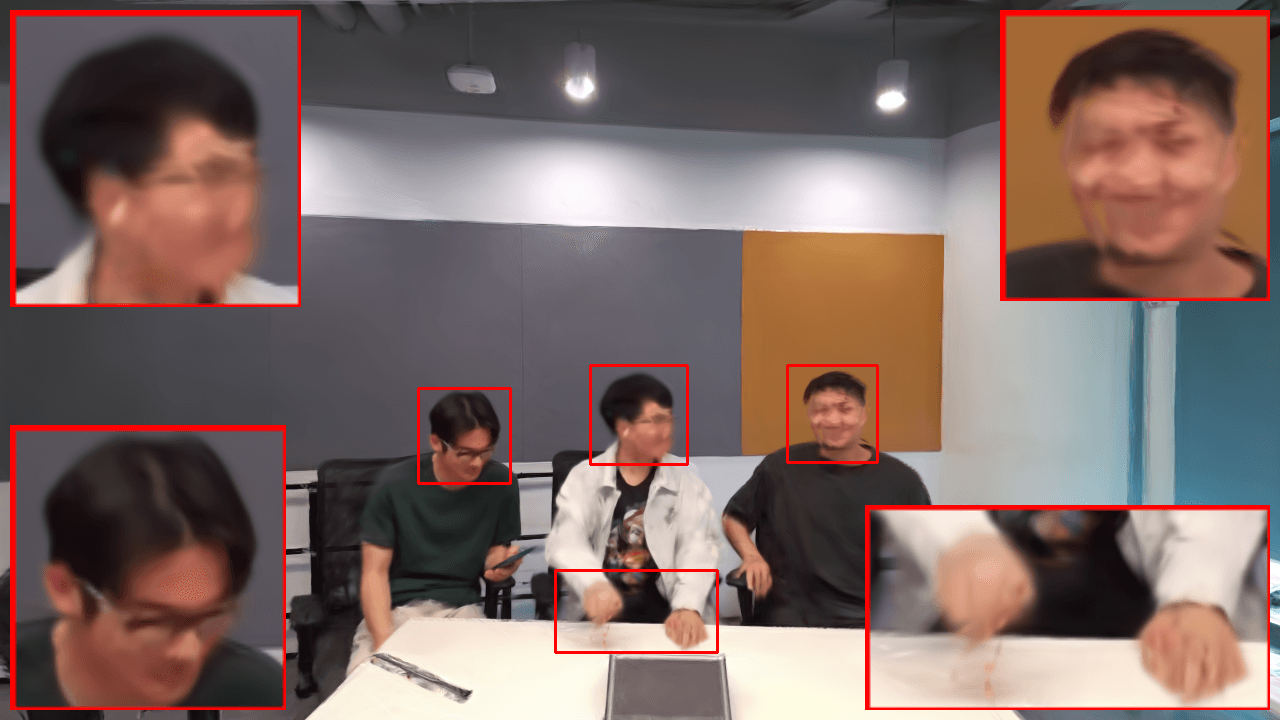}
    \end{subfigure}
    \begin{subfigure}{0.16\textwidth}
        \includegraphics[width=\linewidth]{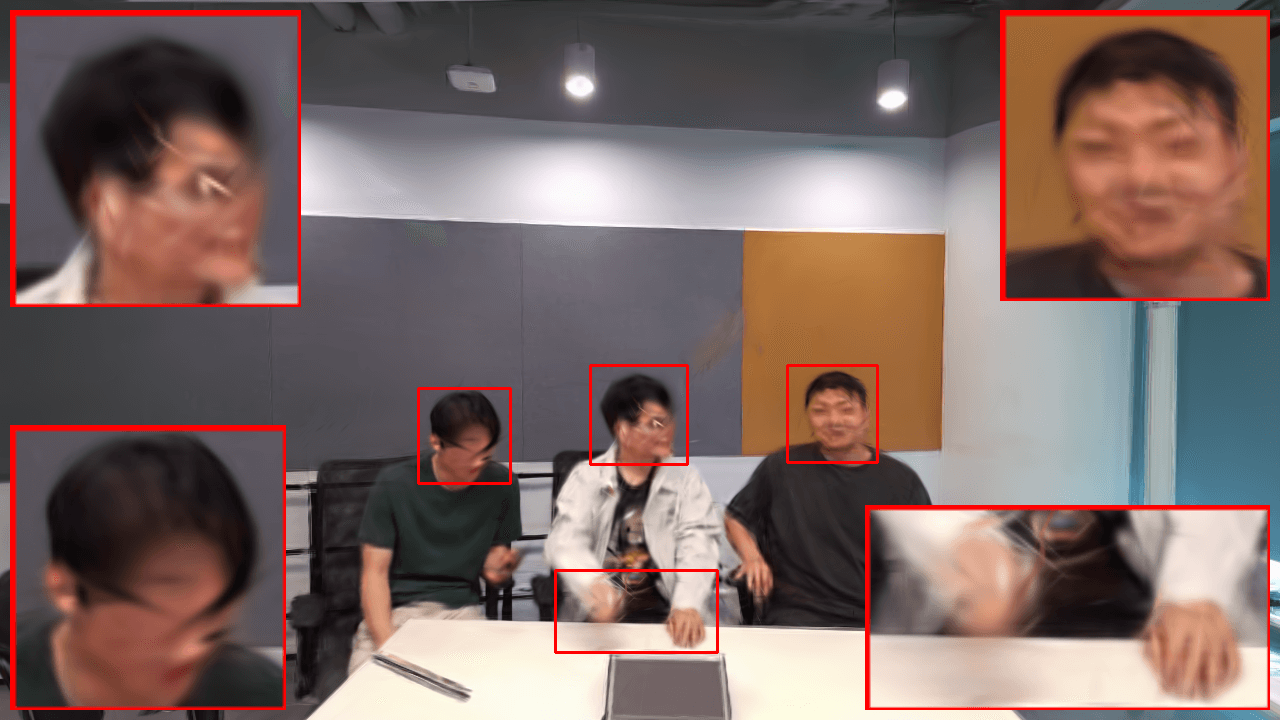}
    \end{subfigure}
    \begin{subfigure}{0.16\textwidth}
        \includegraphics[width=\linewidth]{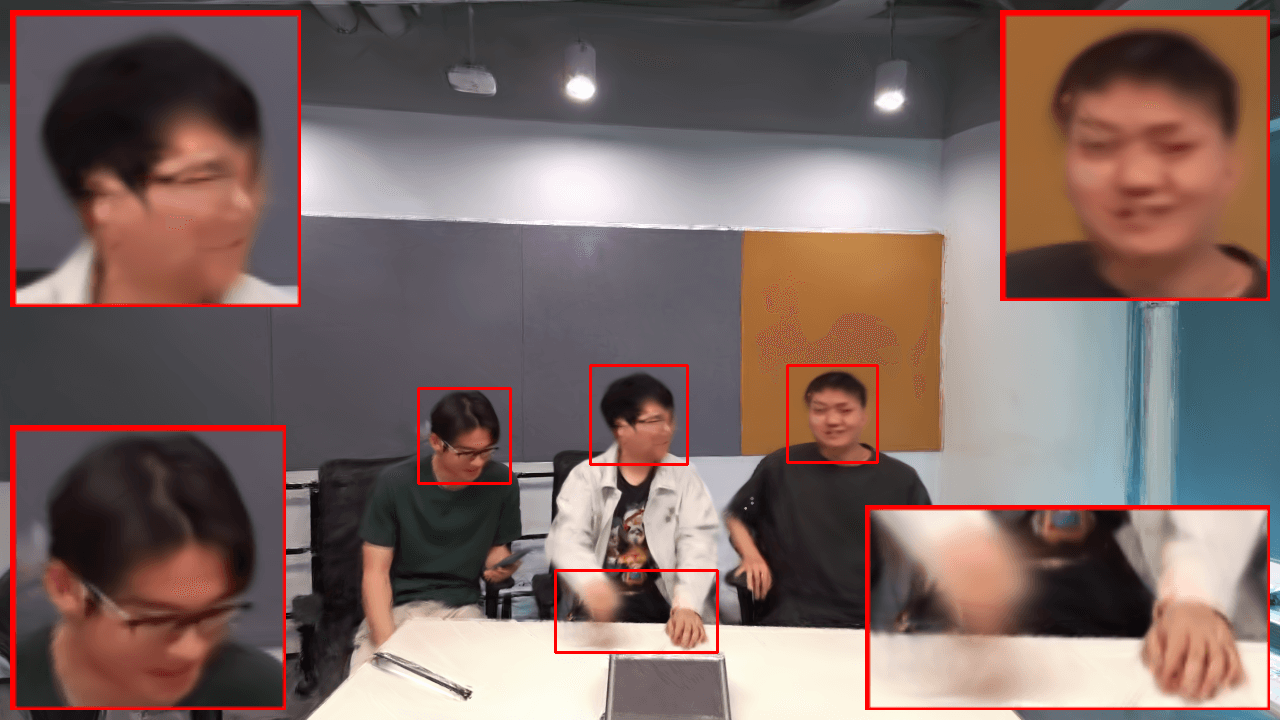}
    \end{subfigure}
    \begin{subfigure}{0.16\textwidth}
        \includegraphics[width=\linewidth]{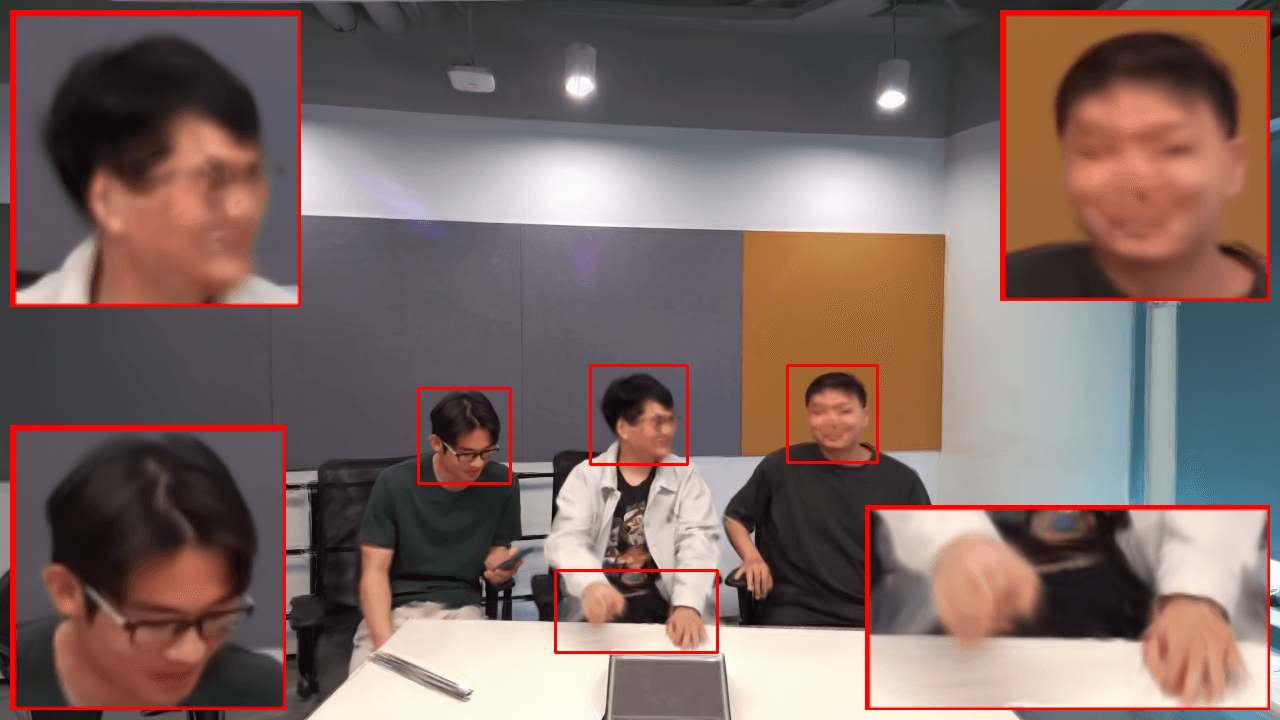}
    \end{subfigure}
    \begin{subfigure}{0.16\textwidth}
        \includegraphics[width=\linewidth]{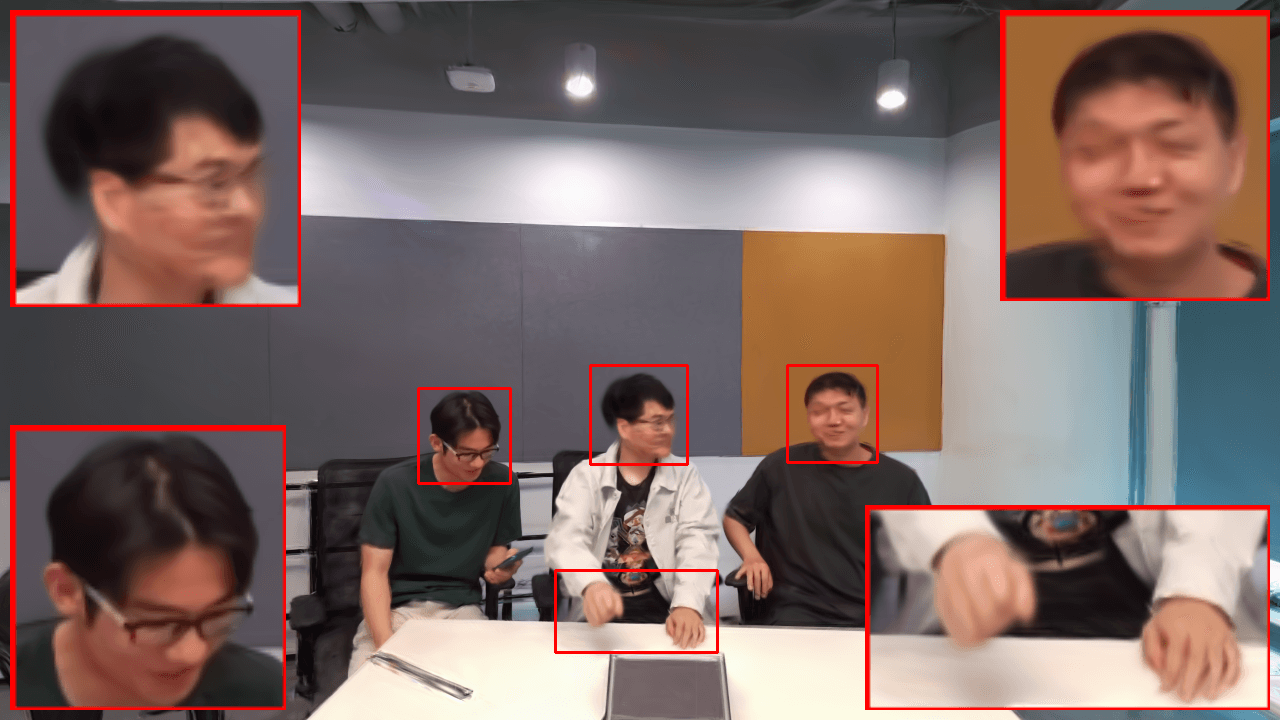}
    \end{subfigure}
    \begin{subfigure}{0.16\textwidth}
        \includegraphics[width=\linewidth]{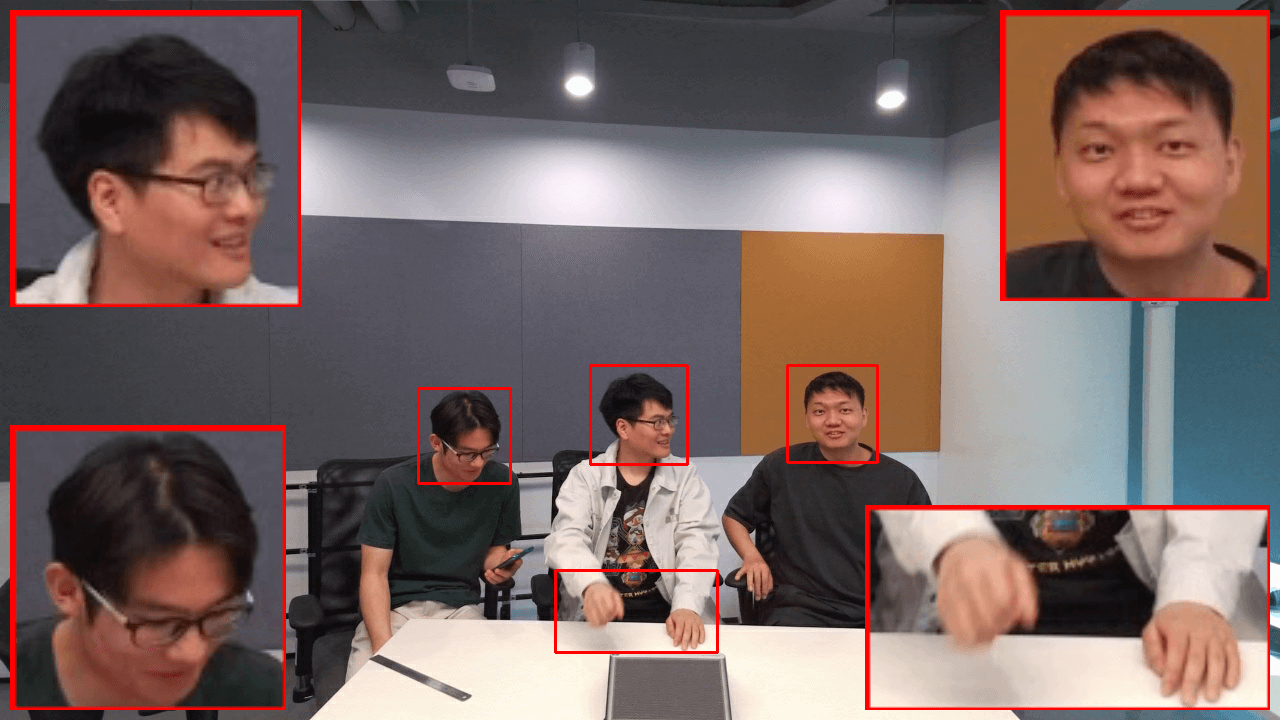}
    \end{subfigure}

    \begin{subfigure}{0.16\textwidth}
      \includegraphics[width=\linewidth]{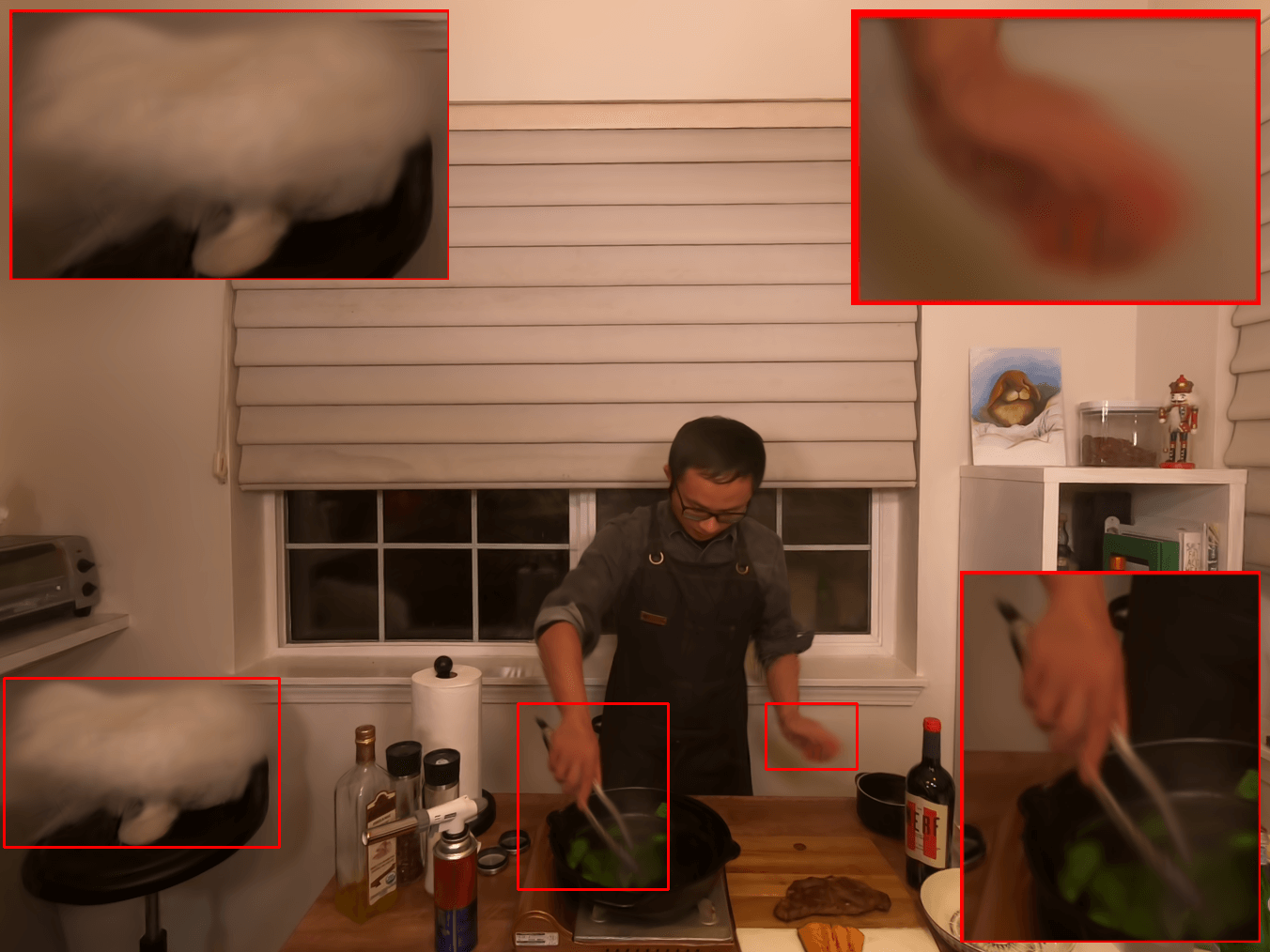}
    \end{subfigure}
    \begin{subfigure}{0.16\textwidth}
        \includegraphics[width=\linewidth]{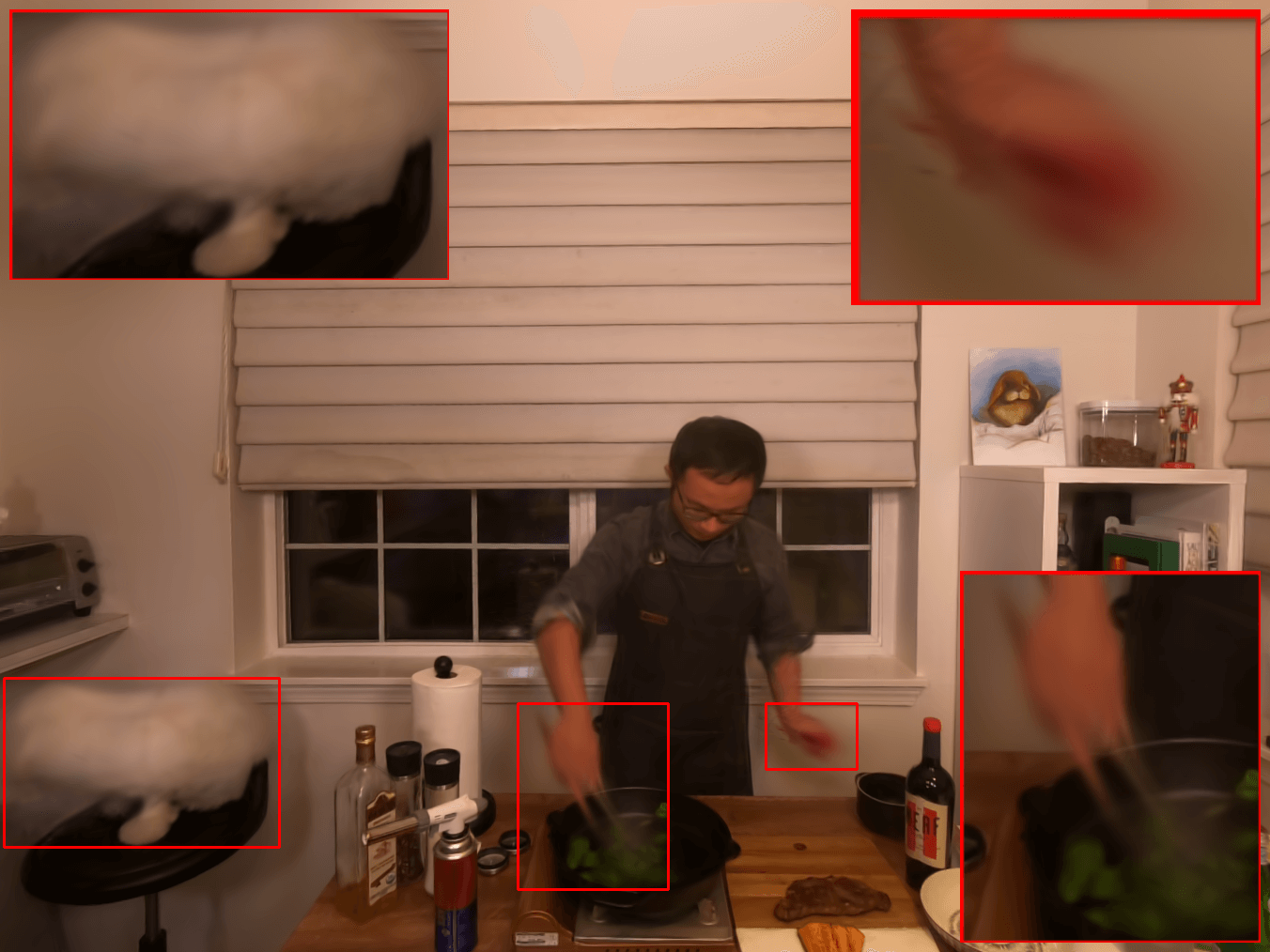}
    \end{subfigure}
    \begin{subfigure}{0.16\textwidth}
        \includegraphics[width=\linewidth]{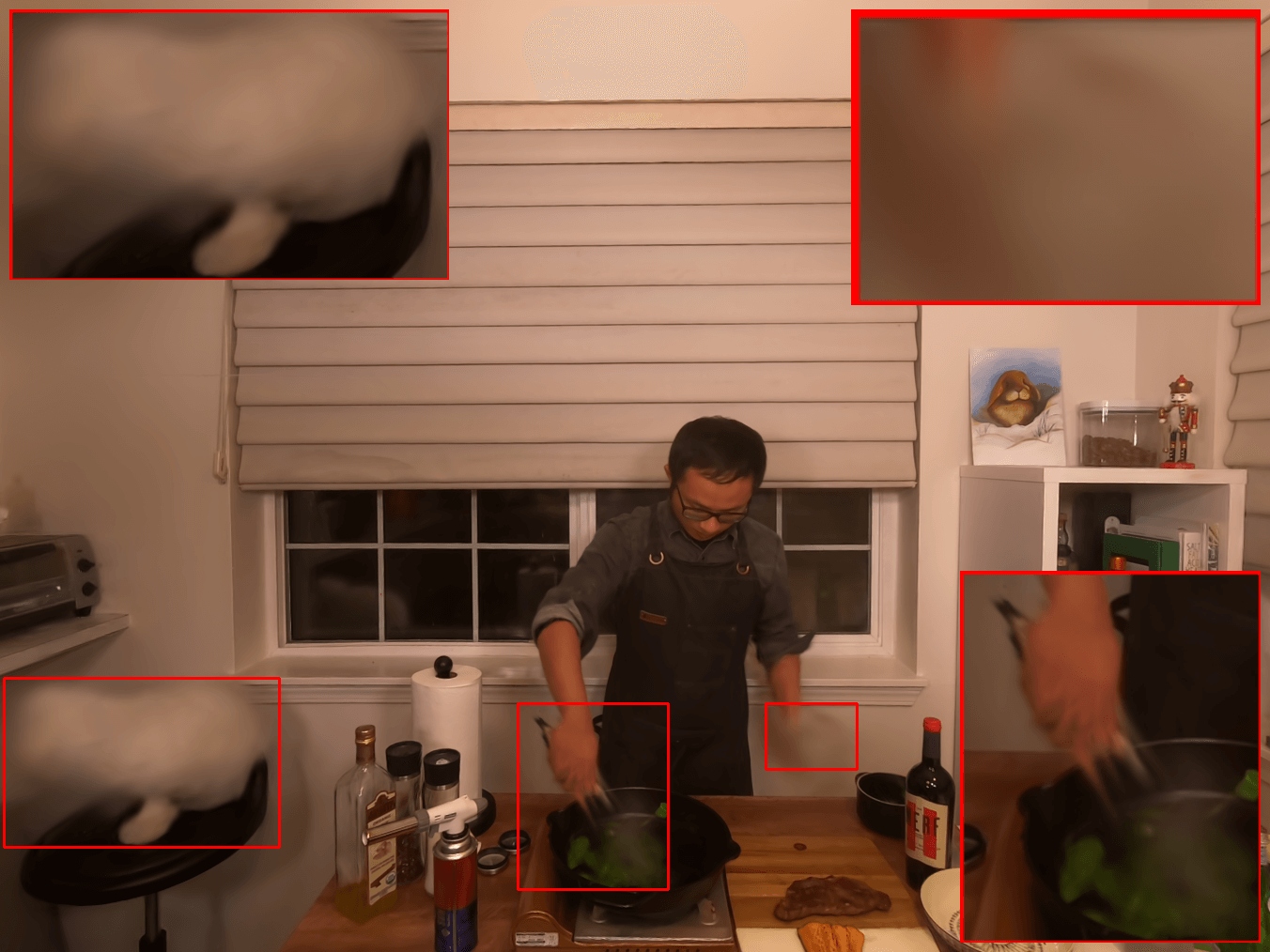}
    \end{subfigure}
    \begin{subfigure}{0.16\textwidth}
        \includegraphics[width=\linewidth]{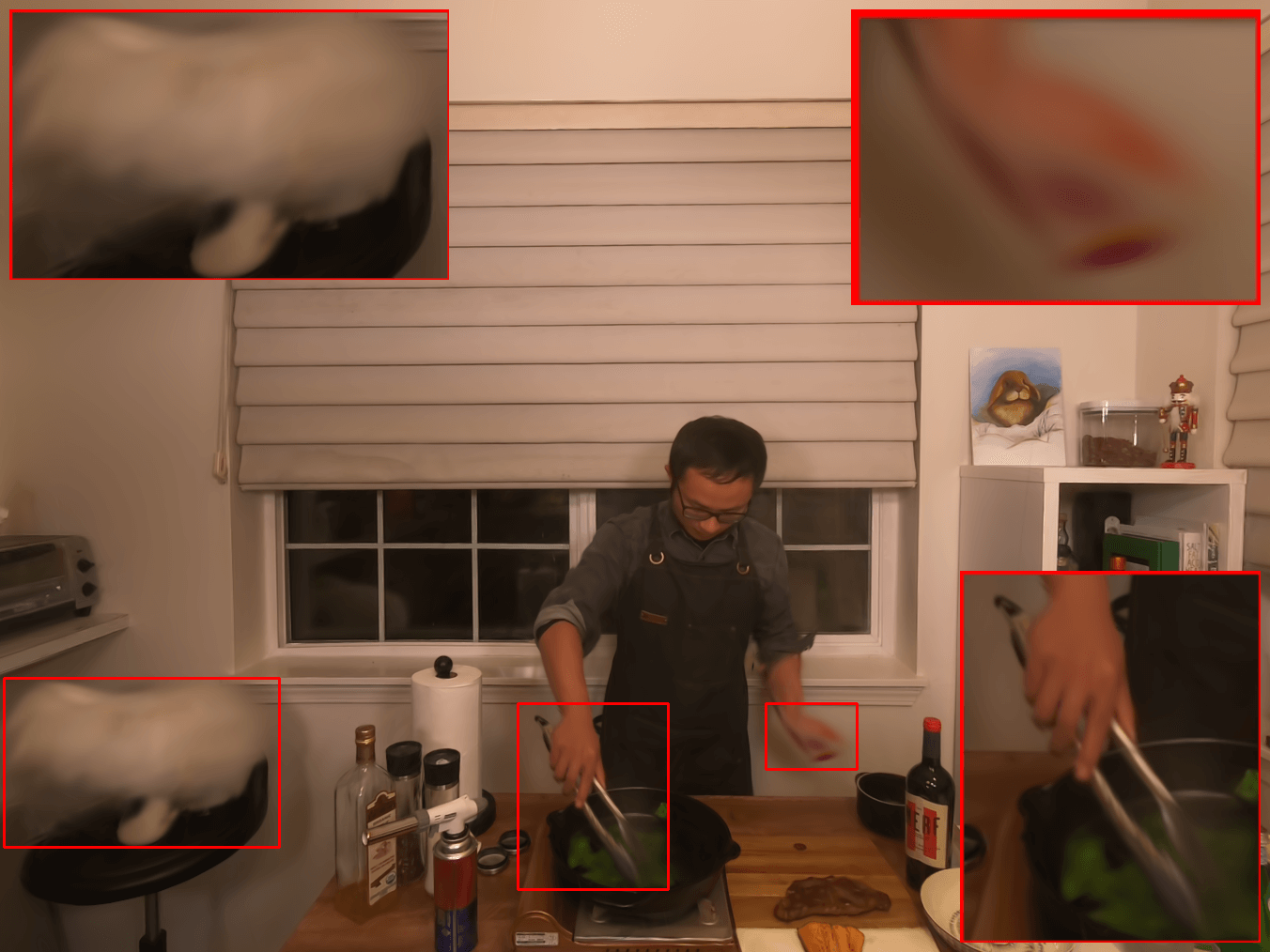}
    \end{subfigure}
    \begin{subfigure}{0.16\textwidth}
        \includegraphics[width=\linewidth]{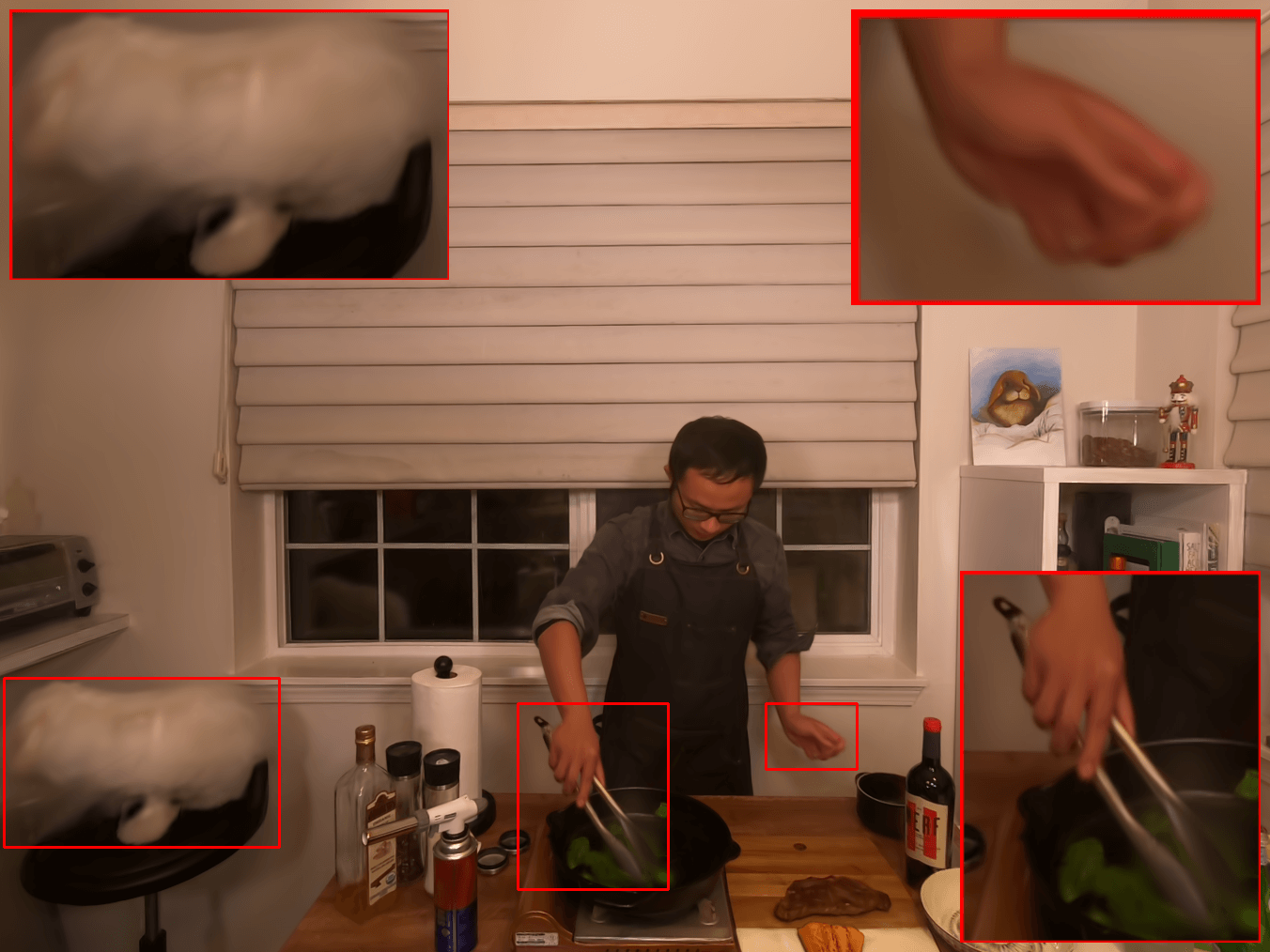}
    \end{subfigure}
    \begin{subfigure}{0.16\textwidth}
        \includegraphics[width=\linewidth]{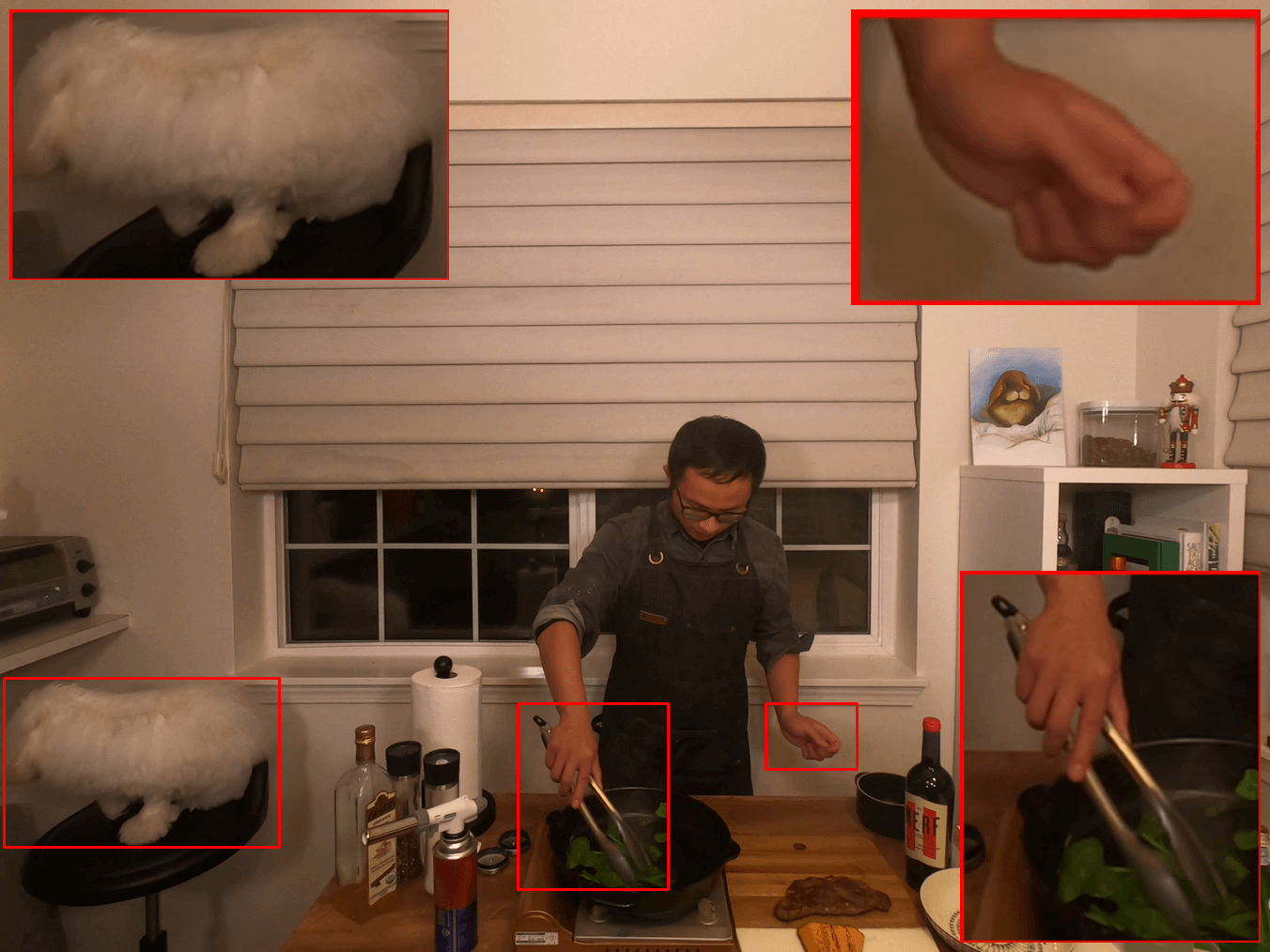}
    \end{subfigure}

    \begin{subfigure}{0.16\textwidth}
      \includegraphics[width=\linewidth]{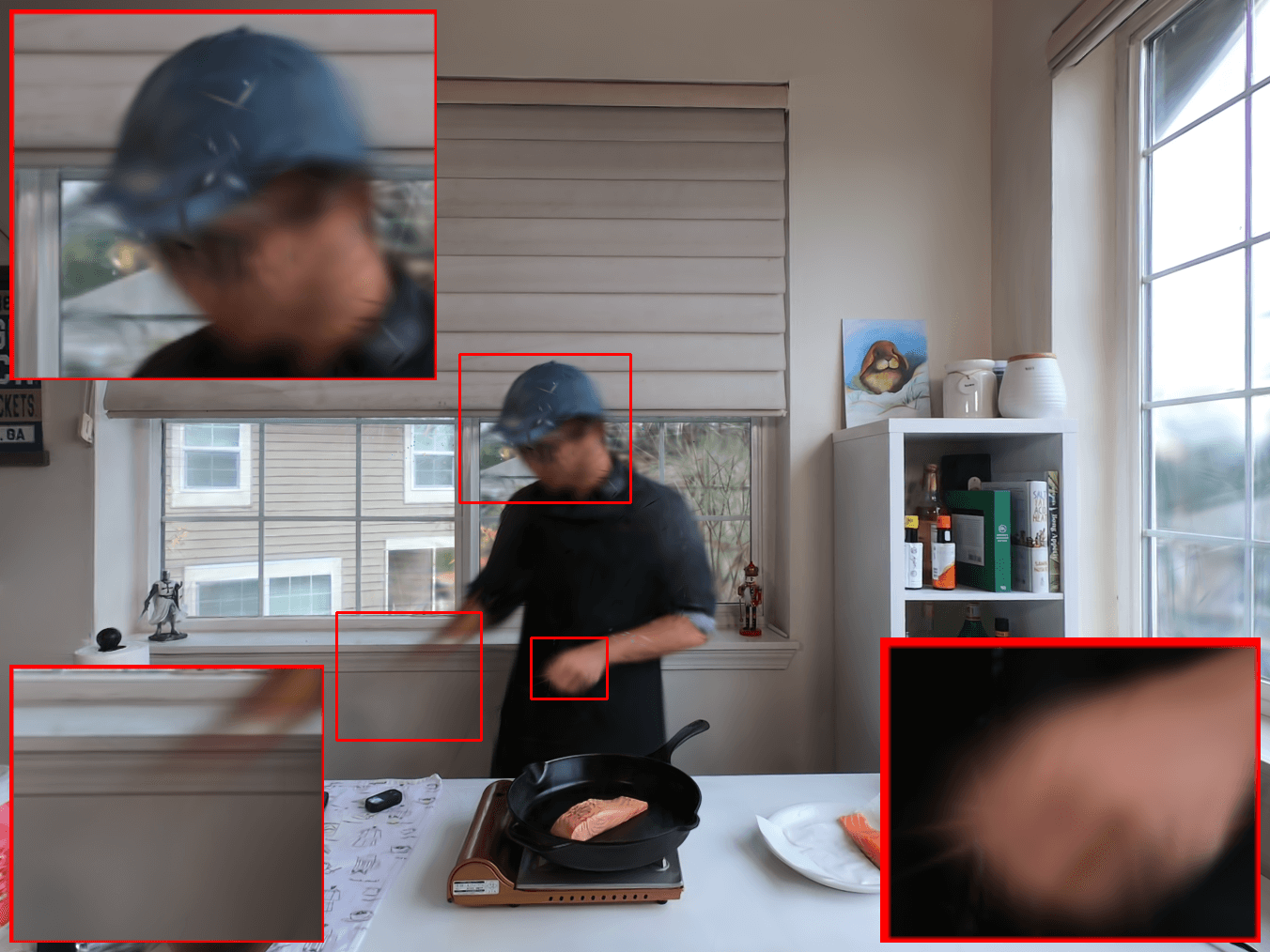}
    \end{subfigure}
    \begin{subfigure}{0.16\textwidth}
        \includegraphics[width=\linewidth]{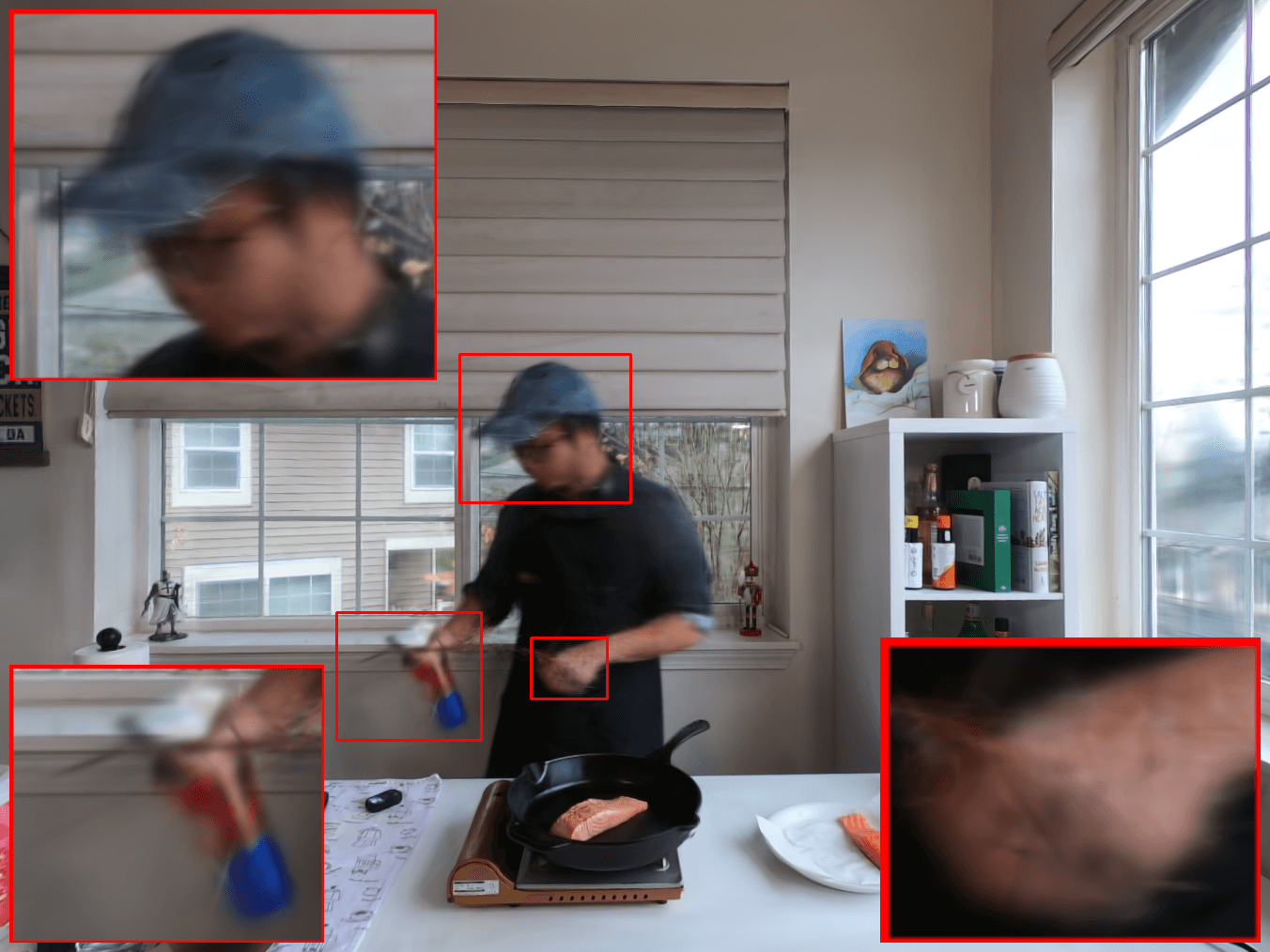}
    \end{subfigure}
    \begin{subfigure}{0.16\textwidth}
        \includegraphics[width=\linewidth]{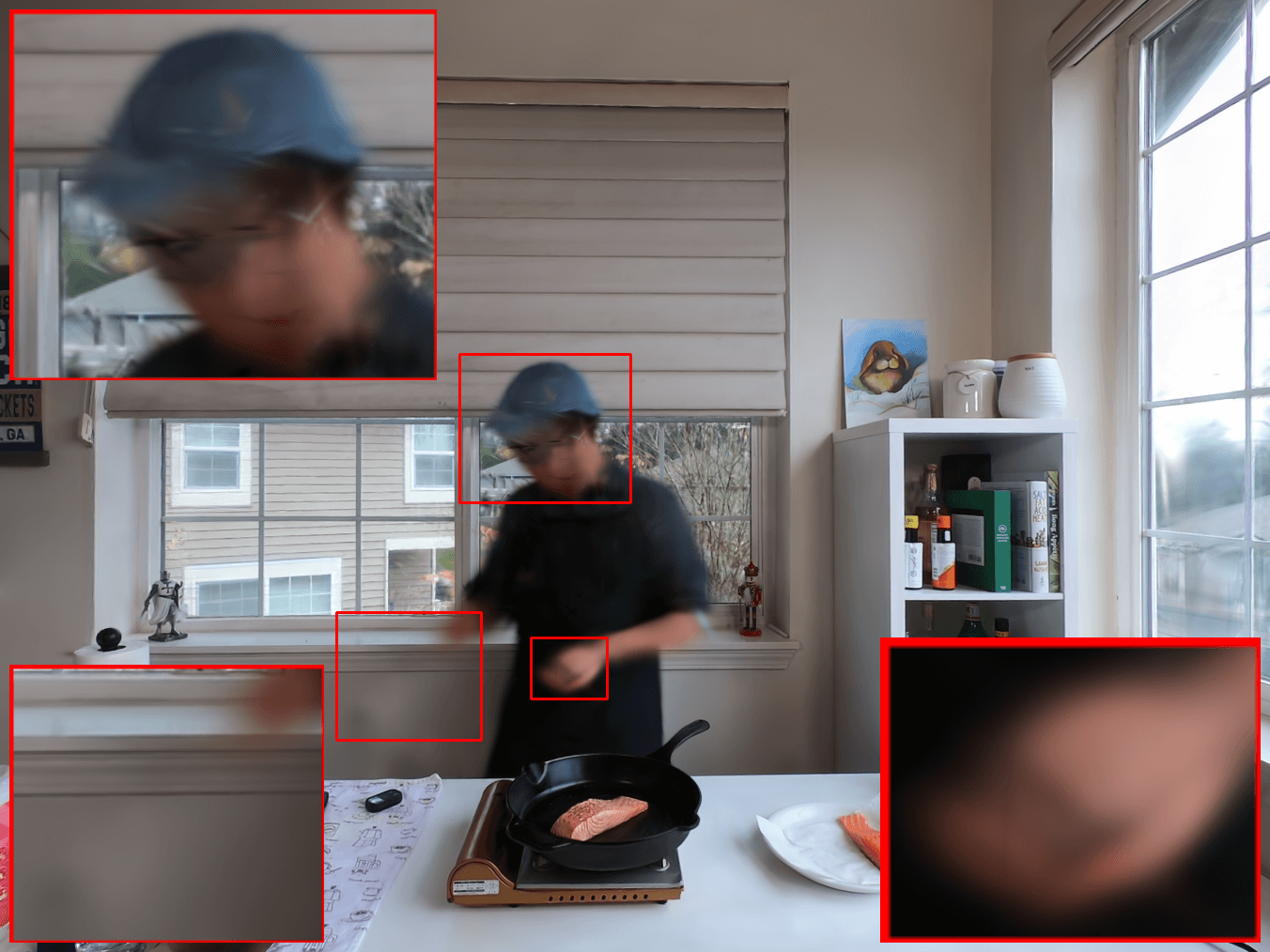}
    \end{subfigure}
    \begin{subfigure}{0.16\textwidth}
        \includegraphics[width=\linewidth]{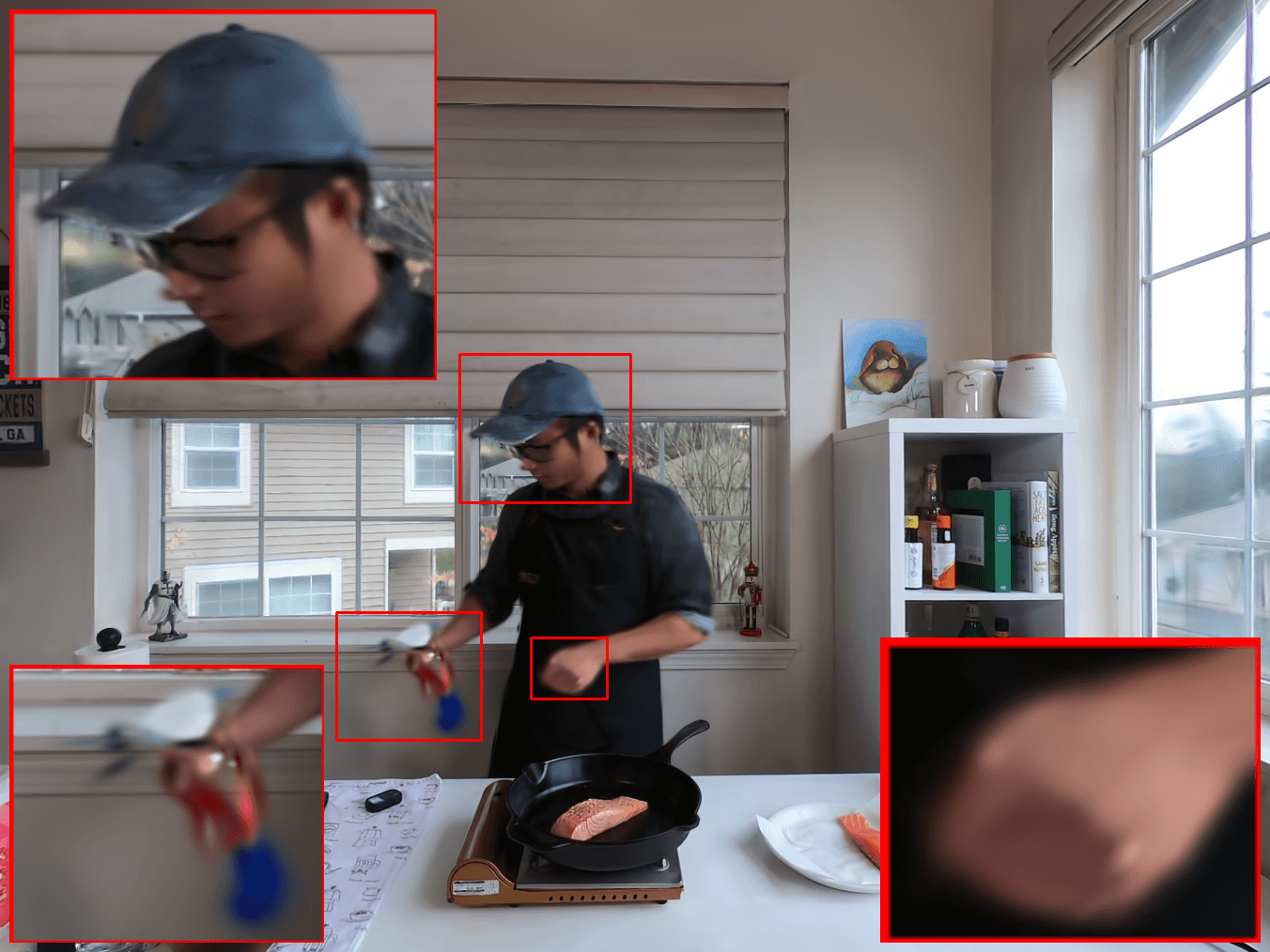}
    \end{subfigure}
    \begin{subfigure}{0.16\textwidth}
        \includegraphics[width=\linewidth]{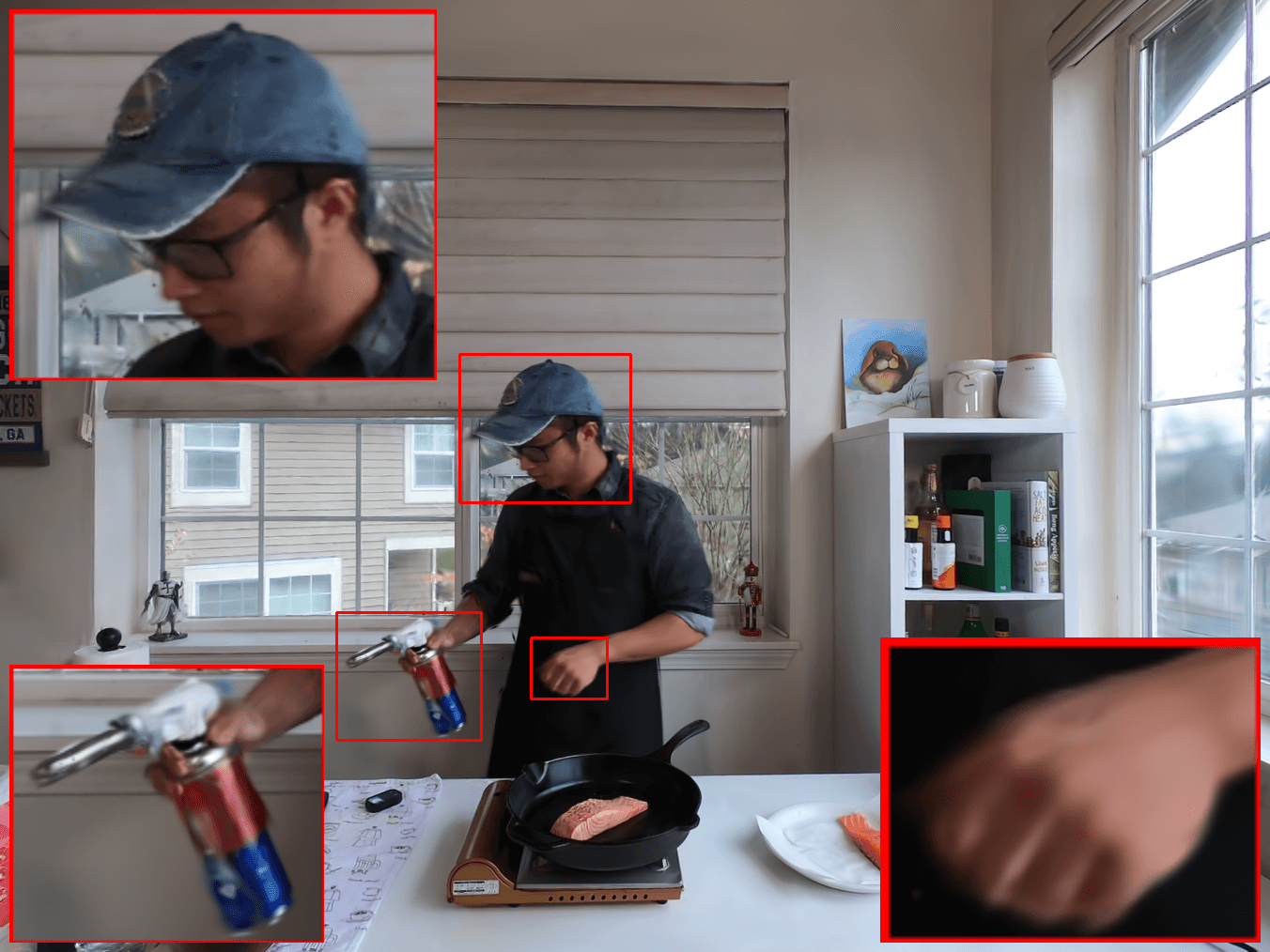}
    \end{subfigure}
    \begin{subfigure}{0.16\textwidth}
        \includegraphics[width=\linewidth]{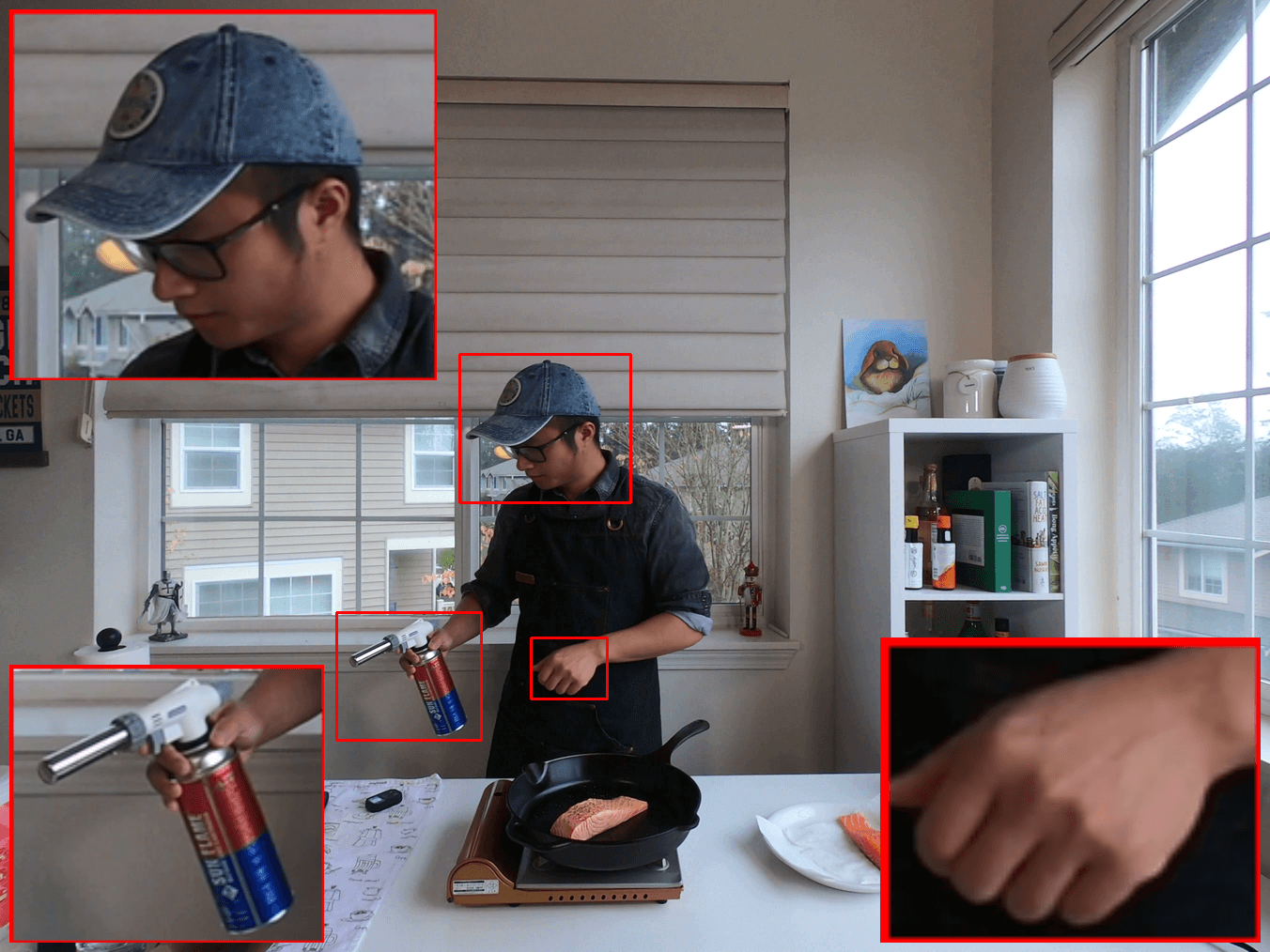}
    \end{subfigure}
    
    \begin{subfigure}{0.16\textwidth}
      \includegraphics[width=\linewidth]{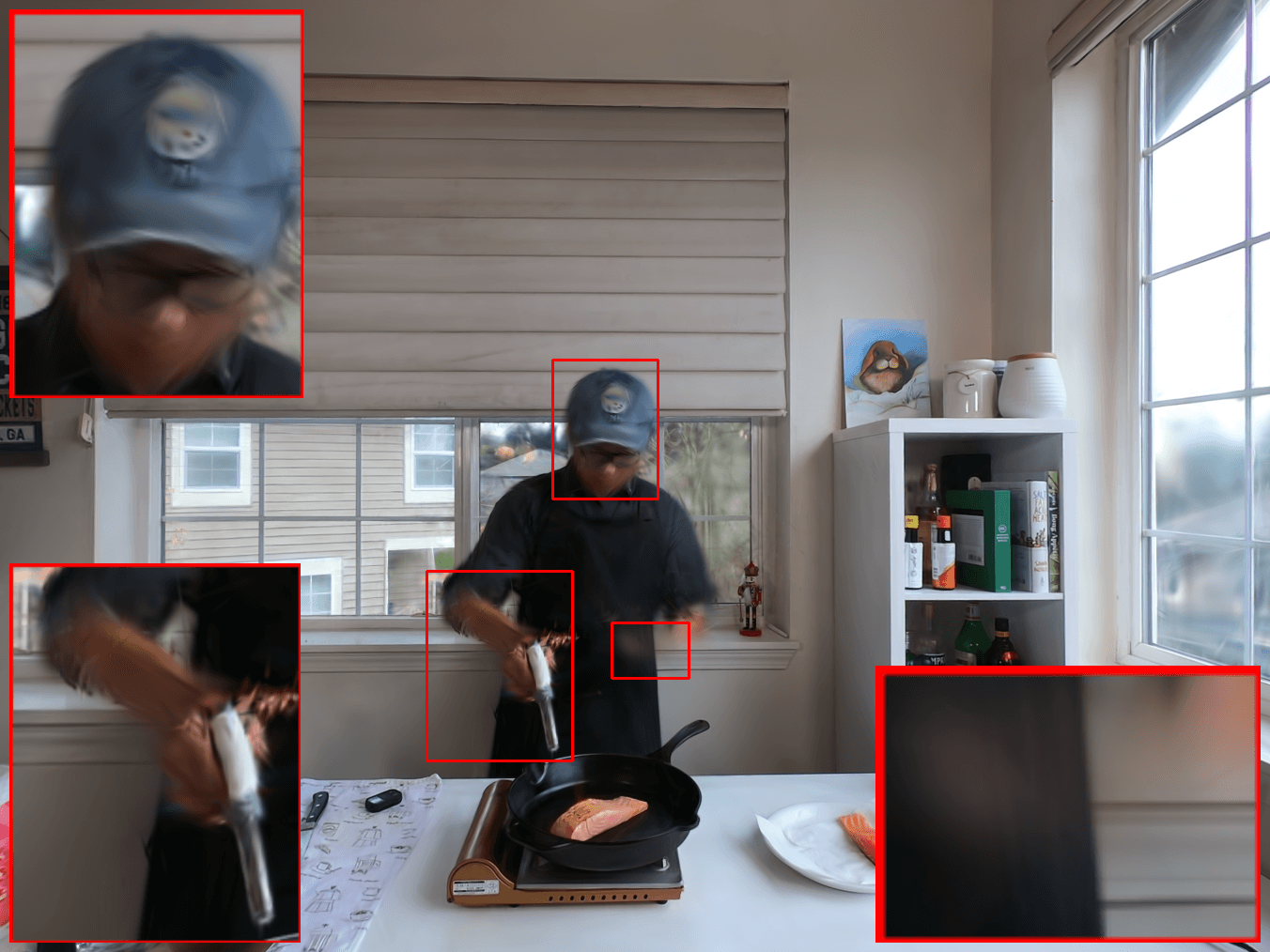}
      \caption*{(a) 4DGaussians}
    \end{subfigure}
    \begin{subfigure}{0.16\textwidth}
        \includegraphics[width=\linewidth]{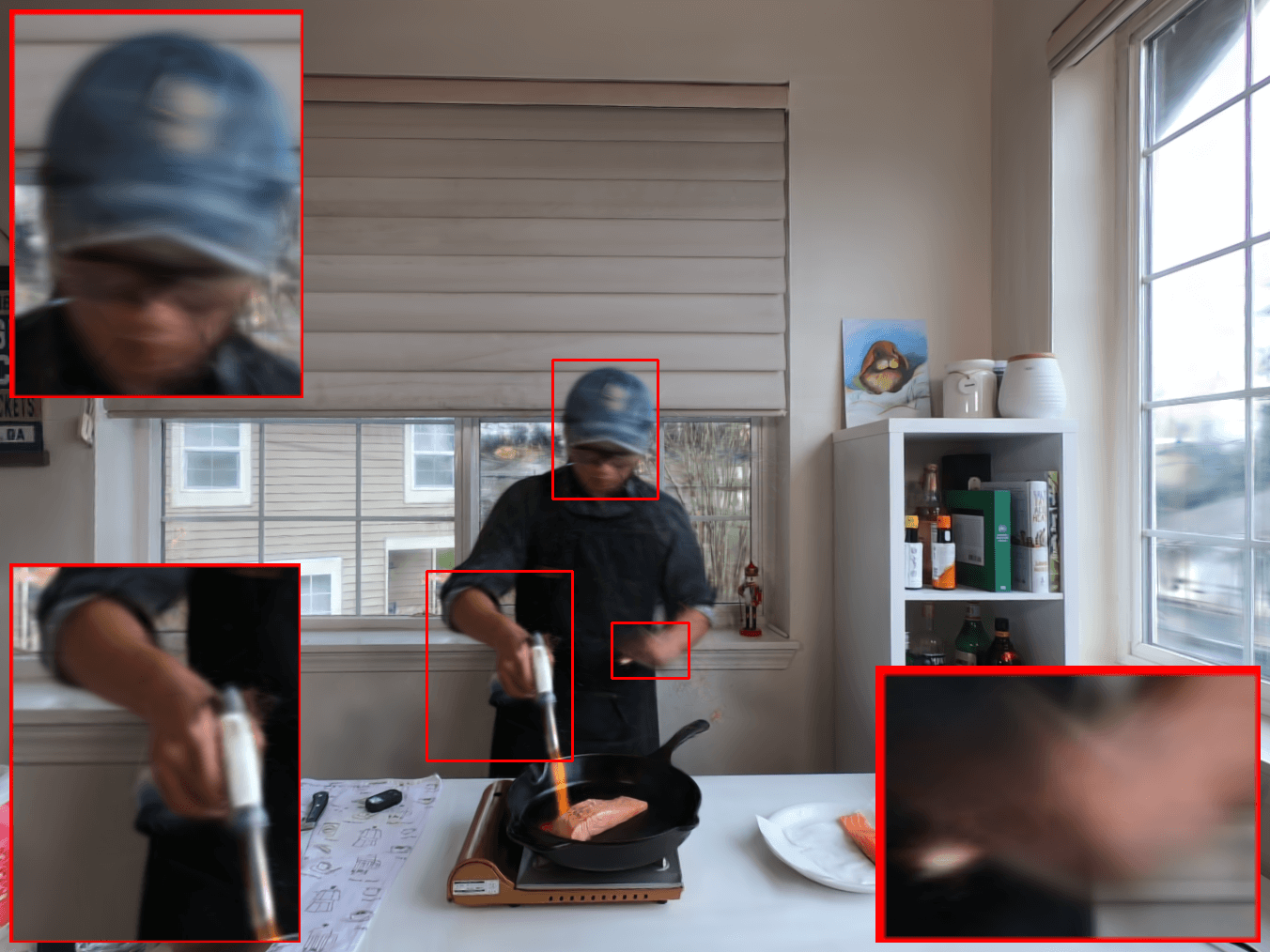}
        \caption*{(b) Ex4DGS}
    \end{subfigure}
    \begin{subfigure}{0.16\textwidth}
        \includegraphics[width=\linewidth]{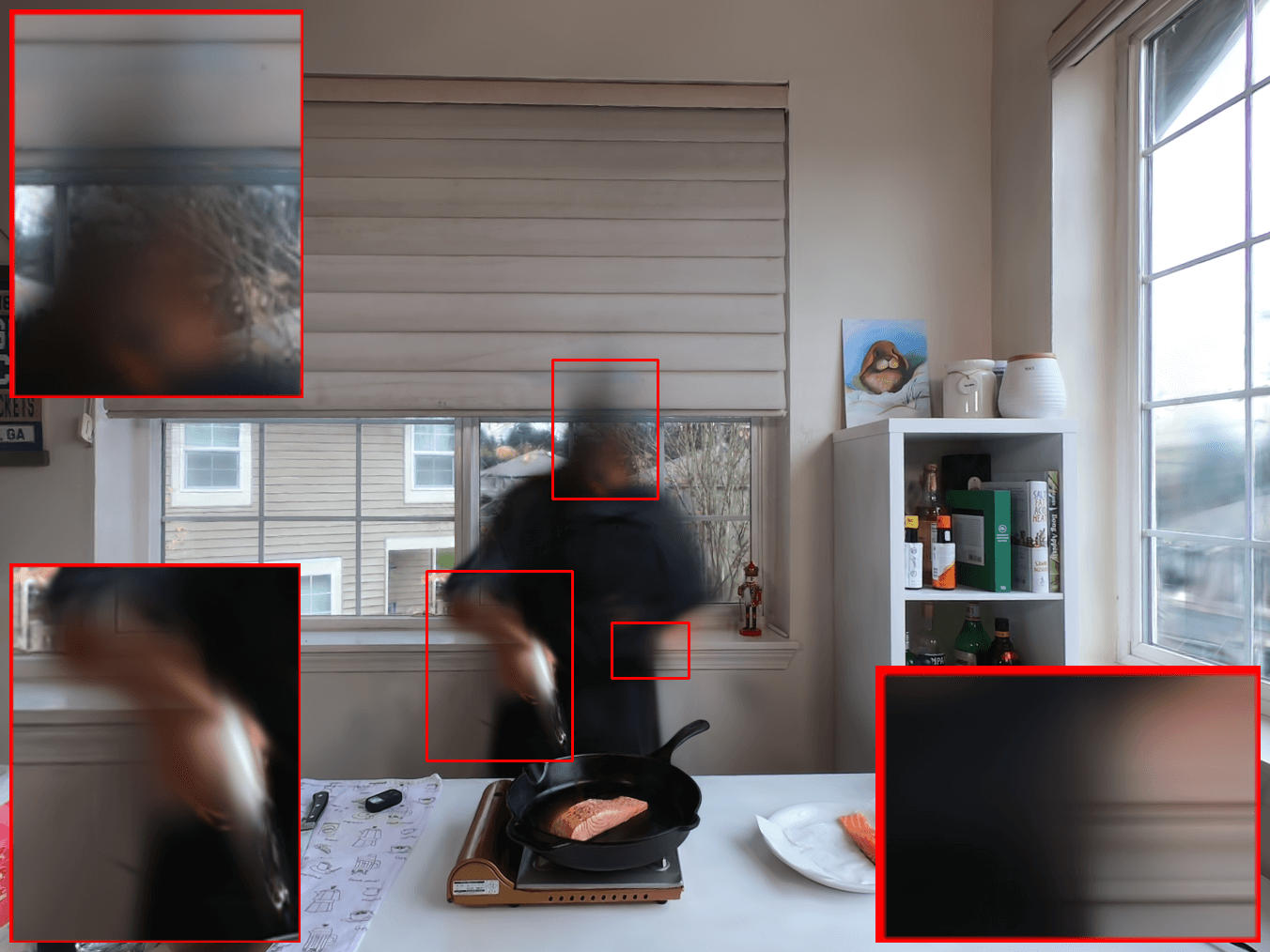}
        \caption*{(c) Swift4D}
    \end{subfigure}
    \begin{subfigure}{0.16\textwidth}
        \includegraphics[width=\linewidth]{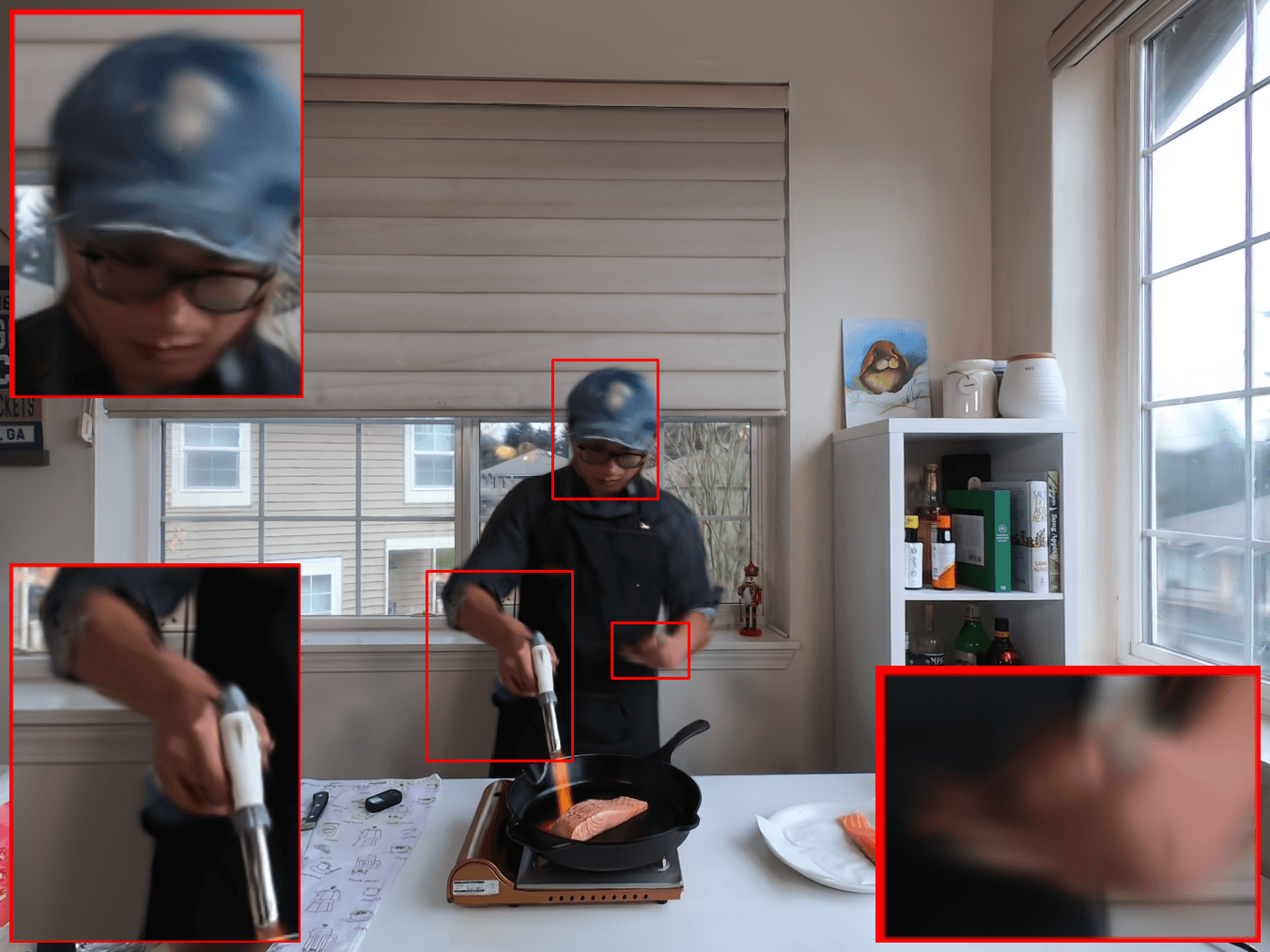}
        \caption*{(d) E-D3DGS}
    \end{subfigure}
    \begin{subfigure}{0.16\textwidth}
        \includegraphics[width=\linewidth]{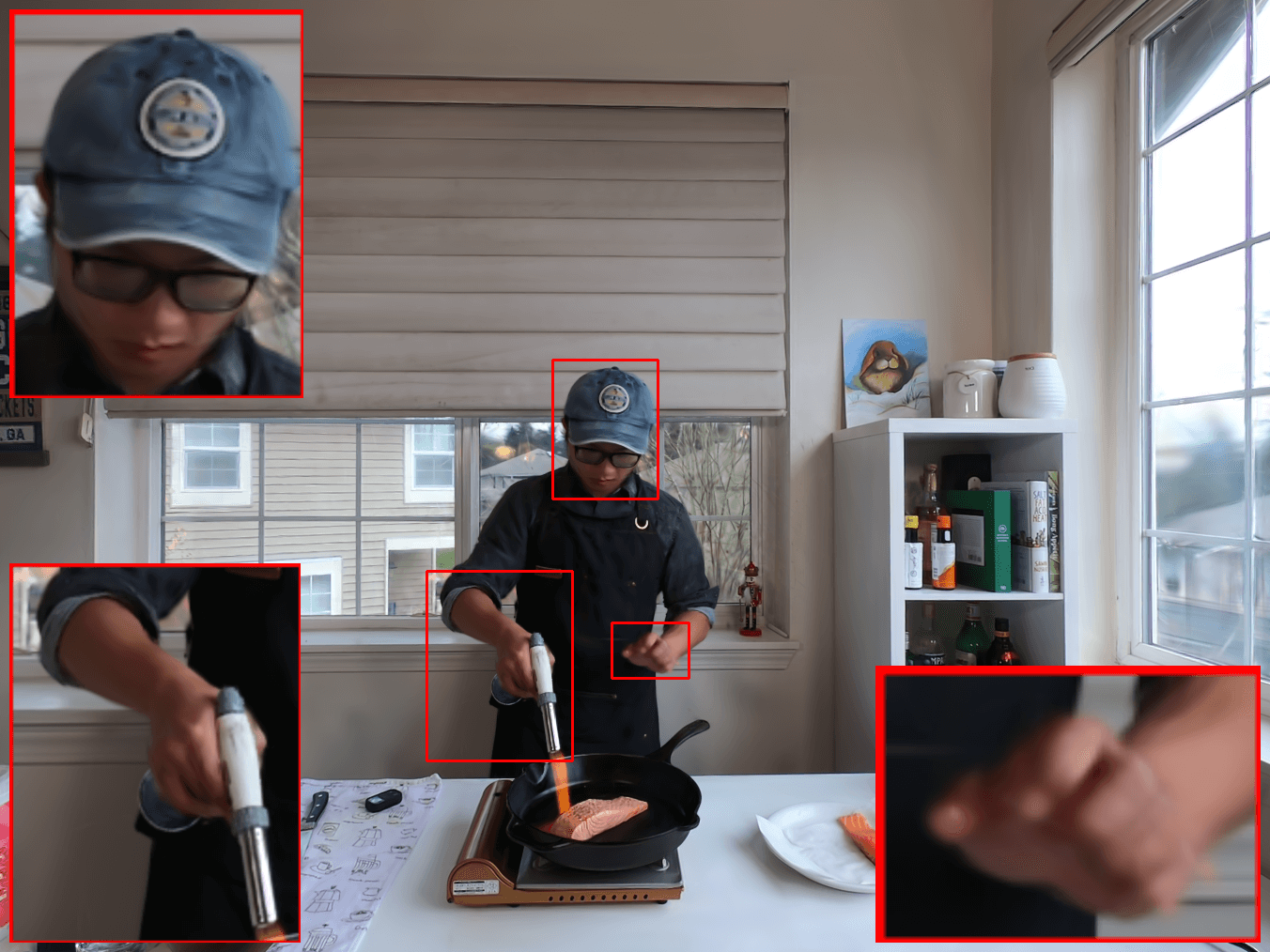}
        \caption*{(e) Ours}
    \end{subfigure}
    \begin{subfigure}{0.16\textwidth}
        \includegraphics[width=\linewidth]{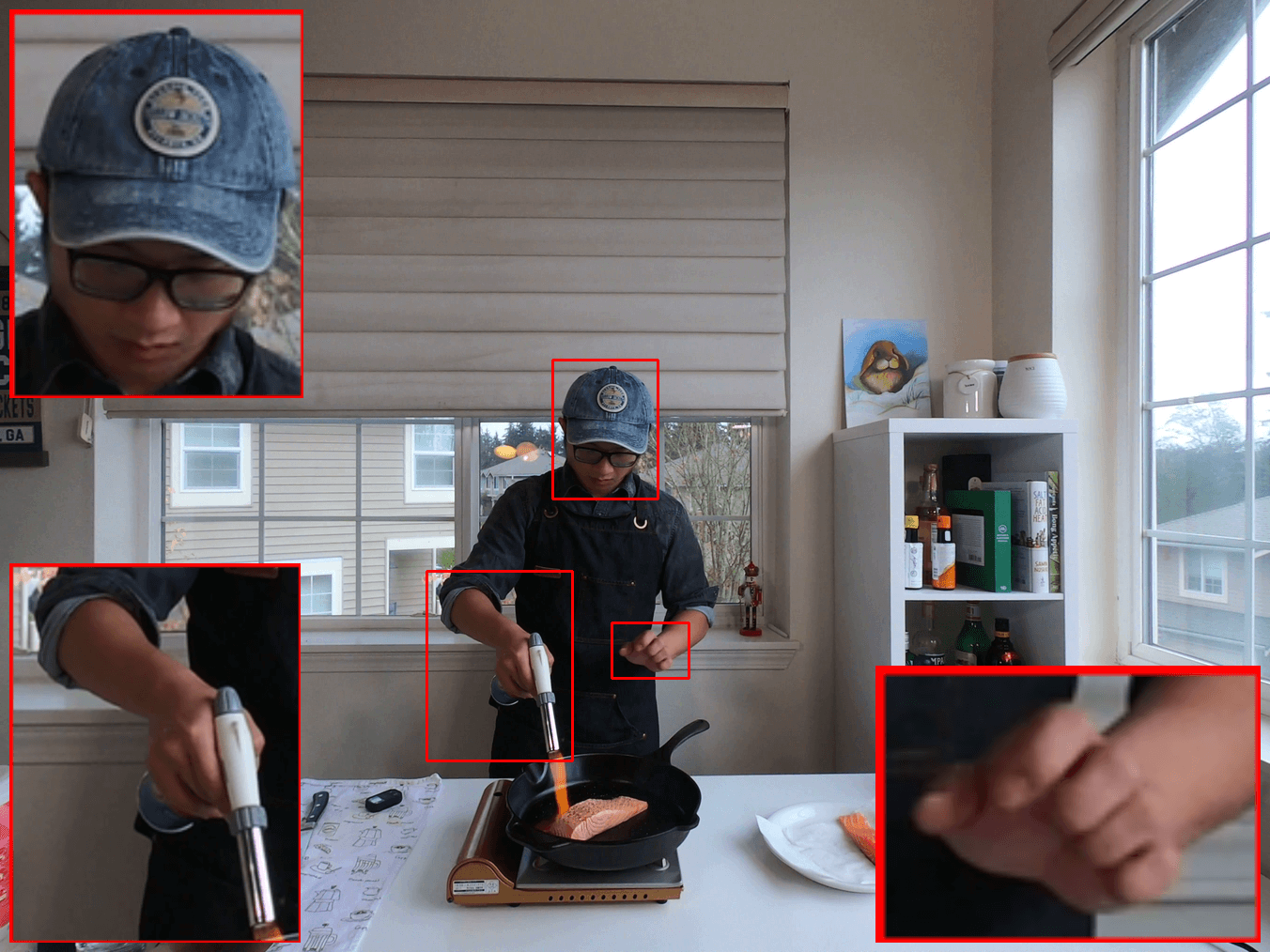}
        \caption*{(f) Ground Truth}
    \end{subfigure}

    \vspace{-2mm}
    \caption{\textbf{Qualitative comparisons against existing SOTA methods on the MeetRoom and N3DV dataset.}}
    \label{fig:visual_Comparisons}
    \vspace{-3mm}
\end{figure*}

\subsection{Cross-Frame Consistency Loss}
Temporal partitioning, while beneficial for modeling complex motions, can introduce visual discontinuities at the partition boundaries. To ensure temporal smoothness, we introduce the cross-frame consistency loss ${L}_{\text{cross}}$, which consists of two components: ${L}_{\text{current}}$ and ${L}_{\text{gt}}$. The ${L}_{\text{current}}$ evaluates rendering consistency at the partition boundary. It compares two renderings of the same frame at this boundary: 
\begin{equation}
    {L}_{\text{current}} = \left\| I_t(G_t,V) - I_t(G_{t^{'}},V) \right\|_1.
\end{equation}
Here, \(G_t\) denotes the set of all 3DGs used for rendering at a timestamp \(t\). $t'$ is the timestamp from the nearest neighboring temporal segment closest to $t$. \(G_t\) comprises two distinct subsets: static 3DGs and dynamic 3DGs. The dynamic 3DGs active at timestamp \(t\), denoted as $G_t^{d}$, are governed by their corresponding deformation networks $F_t^{d}$.
$I_t(G_t,V)$ denotes the image rendered at timestamp $t$ from viewpoint $V$ using $G_t$. By computing the L1 norm between these two rendered images, ${L}_{\text{current}}$ captures the discrepancy between adjacent temporal segments in rendering the same frame. Optimizing this term helps reduce visual discontinuities, ensuring smoother transitions between temporal segments.

Although ${L}_{\text{current}}$ effectively mitigates visual discontinuities across partition boundaries, we observe that relying on it alone can compromise rendering quality. Since ${L}_{\text{current}}$ only enforces self-consistency between adjacent segments without an external reference, continuous optimization can cause them to converge to a consistent but over-smoothed state, leading to perceptible blurring in dynamic regions. To counteract this and anchor the reconstruction to the ground truth, we introduce ${L}_{\text{gt}}$. This loss is designed to enrich the adjacent segment's 3DGs with valuable spatio-temporal context from the current frame. It achieves this by directly supervising their rendering against the ground-truth image of the current view $V$, which is captured at timestamp $t$:
\begin{equation}
    {L}_{\text{gt}} = \left\| I_t(G_{t^{'}},V) - I^{\text{GT}} \right\|_1,
\end{equation}
where $I^{\text{GT}}$ denotes the corresponding ground-truth image at timestamp $t$. This contextual enrichment forces $G_{t'}$ to learn to represent the sharp details of the current frame, thereby preventing over-smoothing and enhancing overall fidelity.

Finally, the overall cross-frame consistency loss, ${L}_{\text{cross}}$, is defined as a weighted combination of ${L}_{\text{current}}$ and ${L}_{\text{gt}}$:
\begin{equation}
    {L}_{\text{cross}} = 0.5 \cdot {L}_{\text{current}} + {L}_{\text{gt}},
\end{equation}
By optimizing ${L}_{\text{cross}}$, we reduce discontinuities across temporal segments while enhancing rendering quality. We apply ${L}_{\text{cross}}$ only for training views whose frame indices are within $5$ frames of any partition boundary. Fig.~\ref{fig:toy_demo} (d) demonstrates its effectiveness through a simple demo.

\begin{table}[t!]
  \caption{\textbf{Quantitative comparison on the N3DV dataset.}
  $^1$ \textit{flame\_salmon} was trained on only frag1.
    $^2$ only reported results on the \textit{flame\_salmon\_frag1} and was trained on 8 GPUs. $^3$ trained with 90 frames. $^4$ trained with 50 frames. Storage, training time, and FPS are measured on \textit{flame\_salmon\_frag1}.
  }
  \label{tab:N3DV_Comparisons_Avg}
  \vspace{-2mm}
  \centering
  \resizebox{\columnwidth}{!}{%
    \begin{tabular}{@{}l|cccccc@{}}
      \toprule
      Method & PSNR$\uparrow$ & SSIM$\uparrow$ & LPIPS$\downarrow$ & Storage$\downarrow$ & Training Time$\downarrow$ & FPS$\uparrow$ \\
      \midrule
      DyNeRF$^{1,2}$                 & 29.58 & - & 0.083 & 56MB                         & 1344 hours                    & 0.01                  \\
      NeRFPlayer$^{1,3}$              & 30.69           & 0.932 & 0.111 & 1654MB                        & 5 hours 36 mins                    & 0.06                  \\
      Mix Voxels              & 30.30           & 0.918 & 0.127 & 512 MB                         & 1 hour 28 mins                    & 1.01                  \\
      K-Planes                & \cellcolor{second}30.86           & \cellcolor{third}0.939 & 0.096 & 309 MB                         & 1 hour 33 mins                    & 0.15                  \\
      HyperReel$^{4}$               & 30.37           & 0.921 & 0.106 & 1362 MB                         & 8 hours 42 mins                    & 1.19                  \\
      \midrule
      D3DGS            & 28.27           & 0.917 & 0.156 & 75MB                        & 2 hours 17 mins                    &  20.29                   \\
      4DGS            & 30.30           & 0.933 & 0.069 & 3.6GB                        & 7 hours 43 mins                    &  54.36                   \\
      4DGaussians      & 30.19           & 0.917 & 0.061 & 53 MB                         & 1 hour 13 mins  & 78.28 \\
      Ex4DGS                & 30.76           & \cellcolor{second}0.939 & 0.056 & 205 MB          & 1 hour 5 mins                   & 51.46                   \\
      Swift4D                & 30.05& 0.931&0.055& 116MB& 48 mins& 138.00\\
      4DGC                & 30.78 & 0.938 & 0.052& 225 MB& 5 hours 44 mins& 124.61 \\
      LocalDyGS               &30.75            &0.933  &0.053  &102 MB              &42 mins   &109.30   \\
      E-D3DGS               & \cellcolor{third}30.79           & 0.934 & \cellcolor{third}0.051 & 73 MB             & 2 hours 41 mins  & 37.51  \\
      E-D3DGS (seg)       & 30.73           & 0.935 & \cellcolor{second}0.049 & 215 MB             & 8 hours 32 mins  & 37.97  \\
      Ours                  & \cellcolor{first}31.33          & \cellcolor{first}0.944 & \cellcolor{first}0.044 & 65 MB & 1 hour 52 mins  & 75.64  \\
      \bottomrule
    \end{tabular}
  }
  \vspace{-2mm}

\end{table}

\begin{table}
  \caption{\textbf{Quantitative comparison on the Meet Room dataset.}
    Storage, training time, and FPS are calculated on \textit{discussion}.
  }
  \vspace{-1mm}
  \label{tab:MeetRoom_Comparisons_Avg}
  \centering
  \resizebox{\columnwidth}{!}{%
    \begin{tabular}{@{}l|cccccc@{}}
      \toprule
      Method & PSNR$\uparrow$ & SSIM$\uparrow$ & LPIPS$\downarrow$ & Storage$\downarrow$ & Training Time$\downarrow$ & FPS$\uparrow$ \\
      \midrule
      D3DGS            & 25.81           & 0.890 & 0.233  & 36 MB                         & 47 mins                    & 42.51                   \\
      4DGS            & 26.12           & 0.896 & 0.080  & 5.4 GB                         & 6 hours 32 mins                    & 70.54                    \\
      4DGaussians      & 26.16           & 0.894 & 0.081 & 51 MB                         & 1 hour 3 mins  & 77.26\\
      Ex4DGS                & \cellcolor{third}26.46           & 0.895 & 0.083 & 123 MB          & 1 hour 6 mins                   & 117.49                  \\
      Swift4D               & 25.51           & 0.882 & 0.085  &76 MB &20 mins   &109.58   \\
      4DGC               & \cellcolor{second}26.56           & \cellcolor{second}0.901 & \cellcolor{second}0.070  &224MB &3 hours 26 mins   &160.59   \\
      LocalDyGS               & 25.85          &0.888  &0.084  &98 MB              &1 hour 7 mins   &130.30   \\
      E-D3DGS               & 26.24           & 0.896 & 0.081 & 28 MB             & 1 hour 36 mins  & 90.26  \\
      E-D3DGS (seg)               & 26.31           &\cellcolor{third}0.900  & \cellcolor{third}0.073 & 89 MB             & 4 hour 3 mins  & 85.20  \\
      Ours                    & \cellcolor{first}26.72           & \cellcolor{first}0.903 & \cellcolor{first}0.066 & 49 MB & 1 hour 19 mins  & 92.21  \\
      \bottomrule
    \end{tabular}
  }
  \vspace{-2mm}

\end{table}
\section{Experiment}
\label{sec:experiment}
\subsection{Dataset and Metrics}
We evaluate our method on two real-world dynamic scene datasets: N3DV~\cite{DyNeRF} and Meet Room~\cite{StreamRF}.
The N3DV dataset includes videos at 30 FPS captured by 20 cameras. Following previous work~\cite{ed3dgs}, we downsample its images to 1352×1014 and segment the longer flame\_salmon sequence into four 10s clips.
The Meet Room dataset consists of videos at a
resolution of 1280×720 and 30 FPS from 13 cameras. 
We report PSNR, SSIM, and LPIPS for rendering quality, alongside computational costs including training time, rendering speed, and storage, which are calculated on an NVIDIA RTX A6000 unless otherwise specified.
\subsection{Implementation Details}
Our implementation builds upon the E-D3DGS codebase. In our experiments, we set the number of recorded historical positions, $m$, to 300 and the maximum partition level to 3. We provide complete implementation details in the appendix, including training specifics and more hyperparameter settings. Furthermore, we include in-depth analyses of our method's mechanics and extensive supplementary experiments in the appendix, covering more baseline comparisons and additional ablation studies.
\subsection{Comparisons}
\subsubsection{Quantitative Comparisons.}
We compare MAPo against current open-source SOTA methods. To ensure fairness, all 3DGS-based baselines are evaluated with identical point cloud initializations.
In addition to these SOTA baselines, we additionally introduce a simple segmentation baseline, E-D3DGS (seg), for comparison to highlight the advantages of our approach. This baseline applies a naive temporal partitioning strategy by uniformly splitting the video sequence into three independent segments and training a separate E-D3DGS model for each.
We defer direct comparisons with methods that lack public codebases or employ significantly different protocols (e.g., SWinGS, ST-GS) to the appendix, where full details on all baselines and experimental setups are provided. As shown in Tab.~\ref{tab:N3DV_Comparisons_Avg} and Tab.~\ref{tab:MeetRoom_Comparisons_Avg}, our method consistently achieves SOTA rendering quality across both datasets while avoiding prohibitive computational overhead, thus offering a compelling balance between high fidelity and practical resource usage.

\subsubsection{Qualitative Comparisons.}
We present qualitative results in Fig.~\ref{fig:visual_Comparisons}. The comparison highlights that baseline methods often produce degraded results in areas with complex or rapid motion. For example, in cases with fast-moving hands or detailed facial expressions, baseline methods exhibit severe motion blur and loss of detail. 
Benefiting from our temporal partitioning strategy, our approach preserves fine details in highly dynamic regions, yielding renderings that are significantly sharper and more faithful to the ground truth.

\begin{table}[t!]
  \caption{\textbf{Progressive component ablation on Meet Room.} Storage, training time, and FPS are calculated on \textit{discussion}.}
  \vspace{-1mm}
  \label{tab:progressive_ablation}
  \centering
    \begin{adjustbox}{width=\columnwidth}
    \begin{tabular}{@{}l|ccccccc@{}}
      \toprule
      \multicolumn{1}{c|}{\textbf{Configuration}} & \textbf{PSNR$\uparrow$} & \textbf{SSIM$\uparrow$} & \textbf{LPIPS$\downarrow$} & \textbf{Storage$\downarrow$} & \textbf{Time$\downarrow$} & \textbf{FPS$\uparrow$} & \textbf{tOF (Avg/Bnd)$\downarrow$} \\
      \midrule
        \textbf{Baseline} & 26.24 & 0.896 & 0.081 & 28 MB & 1h36m & 90.26 & 0.082 / 0.074 \\
        \textbf{Baseline (seg)} & 26.31 & 0.900 & 0.073 & 89 MB & 4h3m & 85.20 & 0.080 / 0.185 \\
      \multicolumn{8}{@{}l}{\small\textbf{+Partition}} \\
      \quad \textbf{Temporal Partition} \\
      \qquad 1.1 +Max Dis & 26.52 & 0.901 & 0.070 & 65 MB & 1h41m & 55.21 & 0.079 / 0.084 \\
      \qquad 1.2 +Var & 26.63 & 0.903 & 0.067 & 67 MB & 1h42m & 54.56 & 0.079 / 0.082 \\
      
      \quad \textbf{Static Partition} \\
      \qquad 2.0 +Static & 26.60 & 0.903 & 0.066 & 48 MB & 1h12m & 92.59 & 0.079 / 0.081 \\
      
      \multicolumn{8}{@{}l}{\small\textbf{+Cross}} \\
      \quad 3.1 +${L}_{\text{current}}$ & 26.49 & 0.899 & 0.071 & 48 MB & 1h18m & 92.88 & 0.078 / 0.074 \\
      \quad 3.2 +${L}_{\text{gt}}$ & 26.72 & 0.903 & 0.066 & 49 MB & 1h19m & 92.21 & 0.078 / 0.072 \\

      \bottomrule
    \end{tabular}
    \end{adjustbox}

\end{table}
\begin{figure}
    \centering
    \begin{subfigure}{0.24\linewidth}
        \centering
        \includegraphics[width=\linewidth]{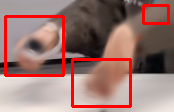}
        \caption*{Baseline}
    \end{subfigure}
    \begin{subfigure}{0.24\linewidth}
        \centering
        \includegraphics[width=\linewidth]{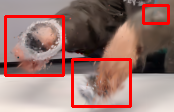}
        \caption*{(1.1) + Max Dis}
    \end{subfigure}
    \begin{subfigure}{0.24\linewidth}
        \centering
        \includegraphics[width=\linewidth]{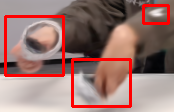}
        \caption*{(1.2) + Var}
    \end{subfigure}
    \begin{subfigure}{0.24\linewidth}
        \centering
        \includegraphics[width=\linewidth]{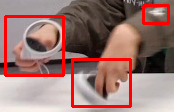}
        \caption*{Ground Truth}
    \end{subfigure}
    \vspace{-1mm}
    \caption{\textbf{Observation of dynamic partition on \textit{Vrheadset}.}}
    \vspace{-2mm}
    \label{fig:DynamicPartition}
\end{figure}
\begin{figure}
    \centering
    \begin{subfigure}{0.32\linewidth}
        \centering
        \includegraphics[width=\linewidth]{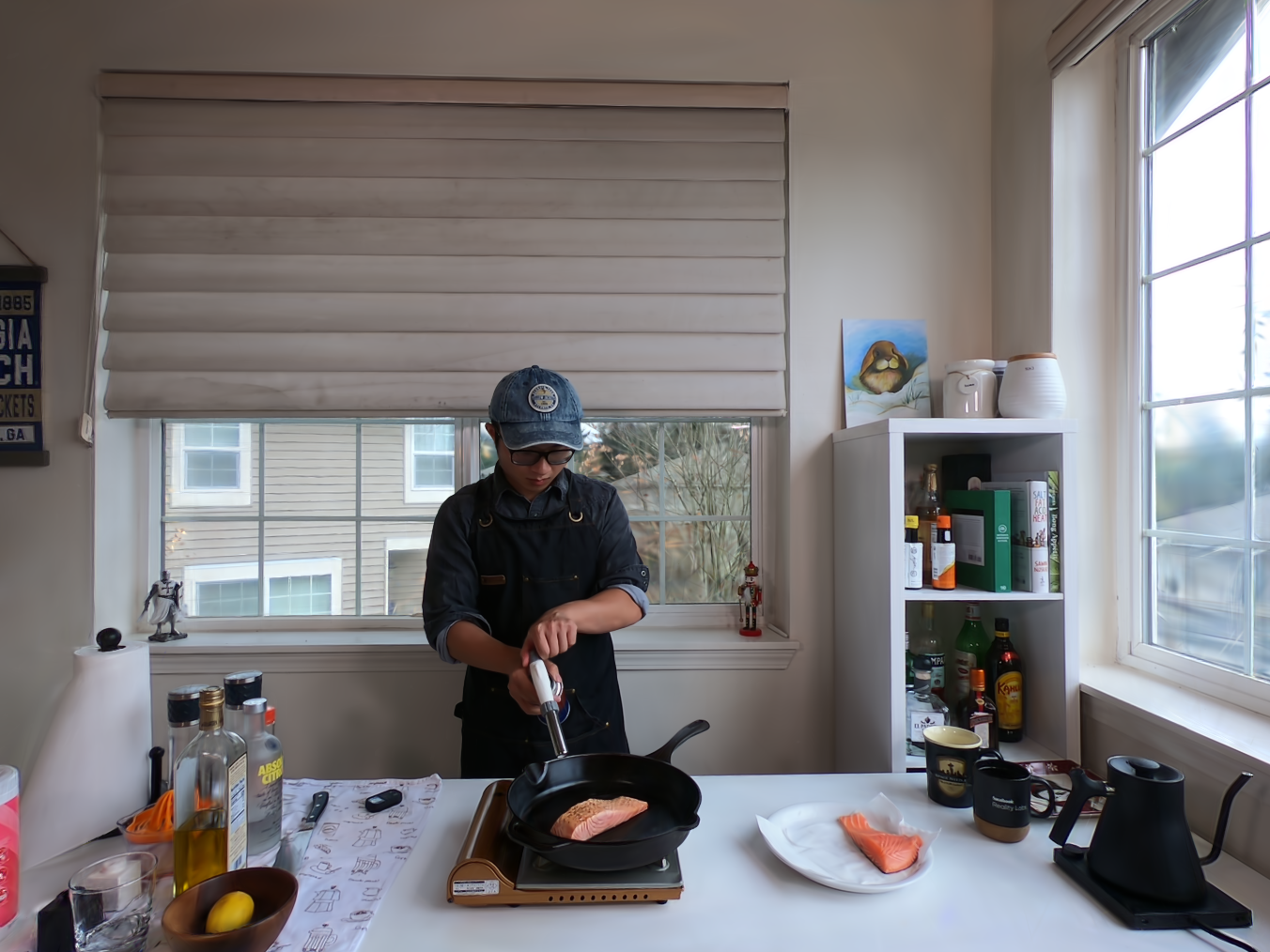}
        \caption*{(1.2) + Var}
    \end{subfigure}
    \begin{subfigure}{0.32\linewidth}
        \centering
        \includegraphics[width=\linewidth]{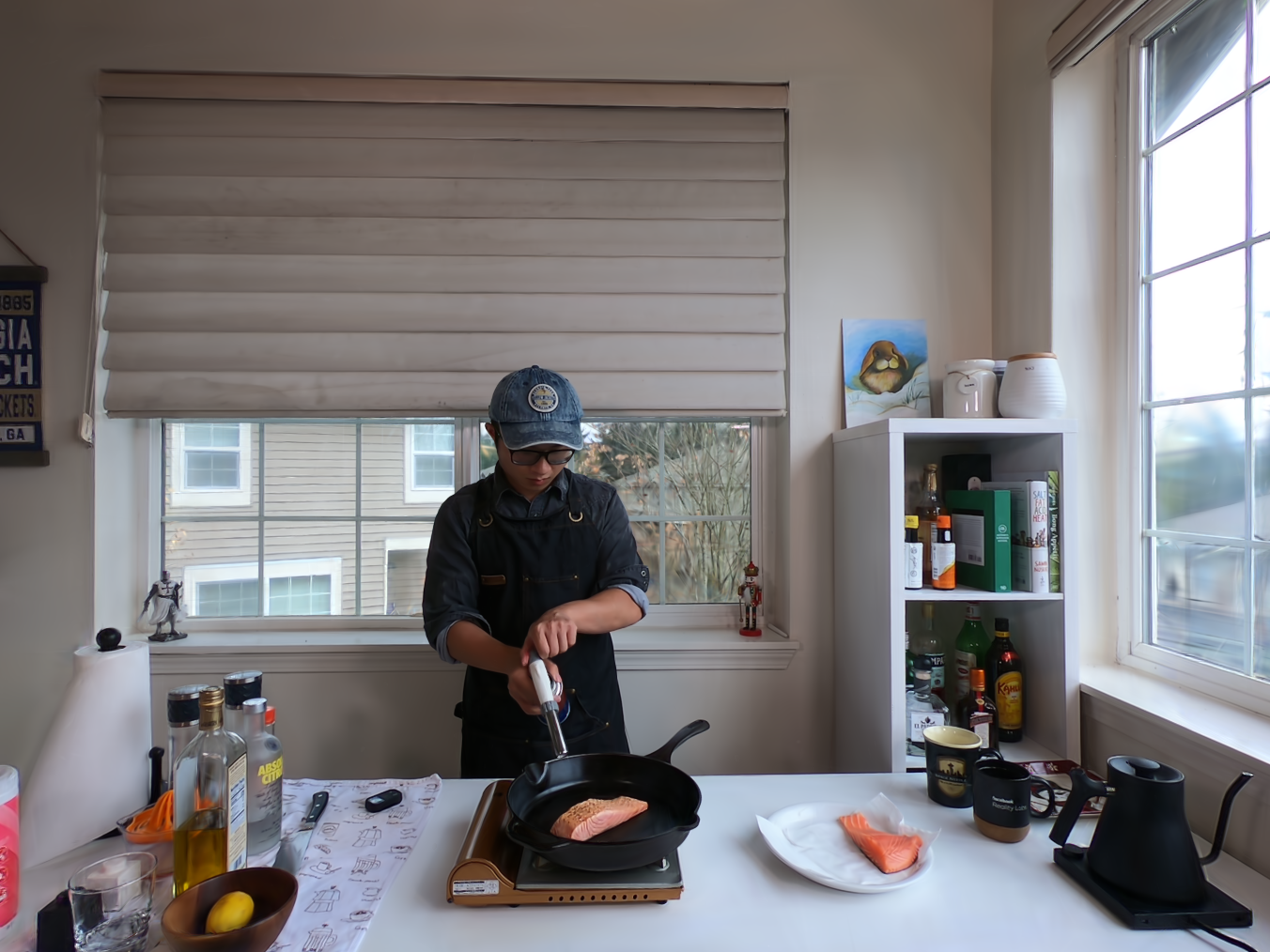}
        \caption*{(2.0) + Static}
    \end{subfigure}
    \begin{subfigure}{0.32\linewidth}
        \centering
        \includegraphics[width=\linewidth]{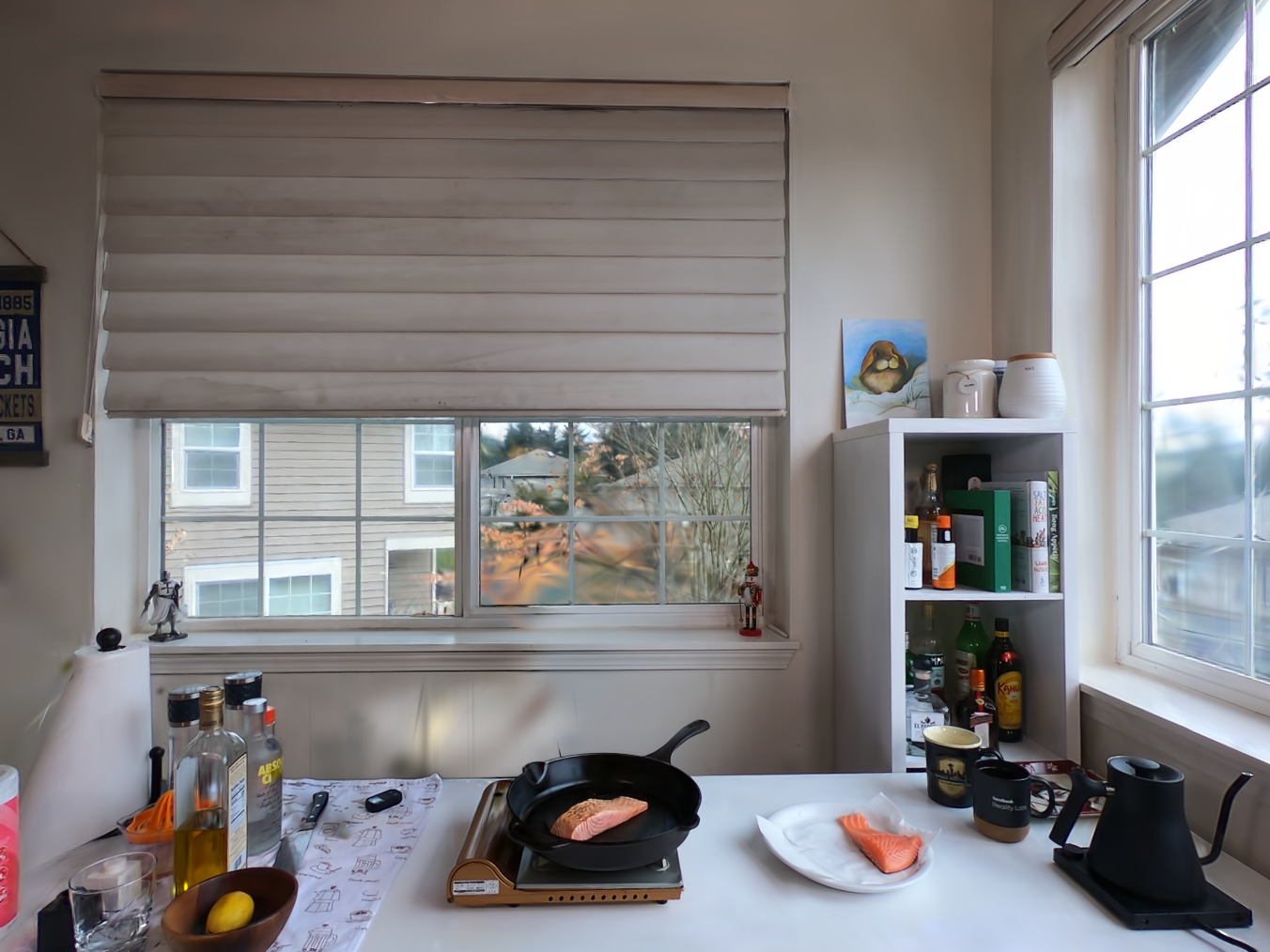}
        \caption*{Static Region}
    \end{subfigure}
    \vspace{-2mm}
    \caption{\textbf{Observation of static partition on \textit{Salmon.}}}
    
    \vspace{-3mm}
    \label{fig:StaticPartition}
\end{figure}
\subsection{Ablation Study and Analysis}
To evaluate our method, we present a progressive ablation study in Tab.~\ref{tab:progressive_ablation}. We establish two reference points: the E-D3DGS model, which serves as our Baseline, and a naive temporal slicing approach of E-D3DGS (Baseline (seg)) for comparison. Building upon the baseline, we incrementally add our proposed components to validate their effect. These components include: 1. \textit{Temporal Partition}, our method for Temporal Partitioning Based on Dynamic Scores, where we use Maximum Displacement (1.1 + Max Dis) for dynamic score computation, then incorporate Variance (1.2 + Var); 2. \textit{Static Partition}, for Static 3D Gaussian Partitioning; and 3. \textit{Cross}, the Cross-Frame Consistency Loss, involving sequential integration of ${L}_{\text{current}}$ (3.1 + ${L}_{\text{current}}$) and ${L}_{\text{gt}}$ (3.2 + ${L}_{\text{gt}}$). Components are added incrementally to evaluate their collective impact. For clarity, we use consistent numbering to denote the variants at each stage.

\begin{figure}[!htbp]
    \centering
    \begin{subfigure}[t]{0.25\linewidth}
        \centering
        \includegraphics[width=\linewidth]{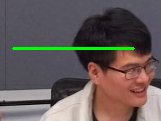}
        \caption*{Annotation}
    \end{subfigure}
    \begin{subfigure}[t]{0.73\linewidth}
        \centering
        \begin{tabular}{cc}
            \begin{subfigure}{0.48\linewidth}
                \centering
                \includegraphics[width=\linewidth]{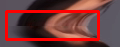}
                \caption*{(2.0) + Static}
            \end{subfigure} &
            \begin{subfigure}{0.48\linewidth}
                \centering
                \includegraphics[width=\linewidth]{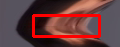}
                \caption*{(3.1) + ${L}_{\text{current}}$}
            \end{subfigure} \\
            \begin{subfigure}{0.48\linewidth}
                \centering
                \includegraphics[width=\linewidth]{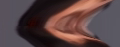}
                \caption*{(3.2) + ${L}_{\text{gt}}$}
            \end{subfigure} &
            \begin{subfigure}{0.48\linewidth}
                \centering
                \includegraphics[width=\linewidth]{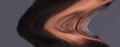}
                \caption*{Ground Truth}
            \end{subfigure}
        \end{tabular}
    \end{subfigure}
    \vspace{-2mm}
    \caption{\textbf{Observation of ${L}_{\text{cross}}$'s effect on visual discontinuity.} The four images on the right are obtained by vertically concatenating the horizontal line in the ``Annotation" over time.}
    \label{fig:CrossFrameJitter}
    \vspace{-2mm}
\end{figure}

\begin{figure}[h]
    \centering
    \centering
    \includegraphics[width=\linewidth]{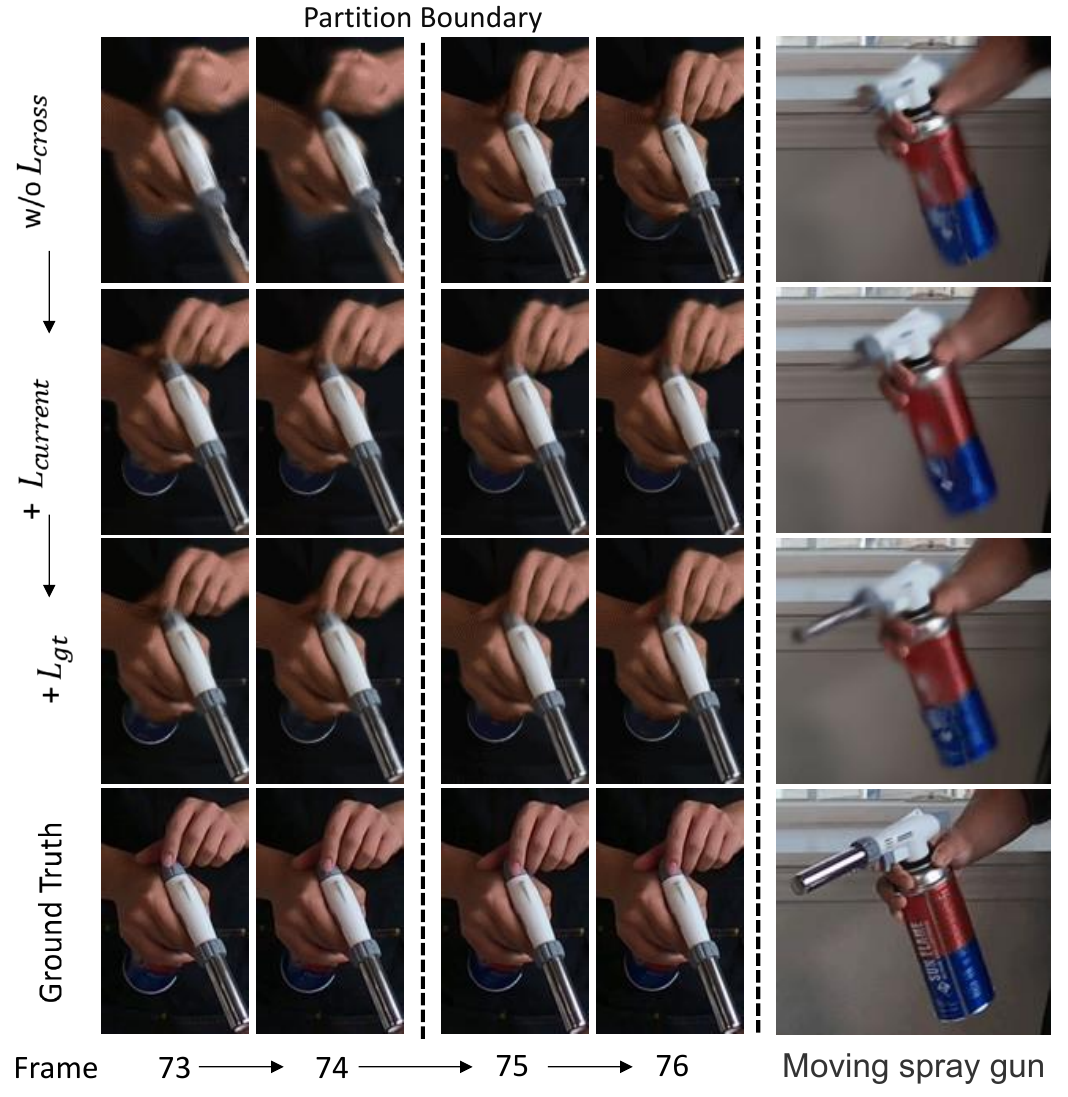}
    \vspace{-2mm}
\caption{\textbf{Ablation study on the $L_{cross}$.} We visualize how $L_{cross}$ improves temporal consistency and rendering quality across a partition boundary (frames 74-75). The first four columns show the frame sequence to evaluate smoothness, while the fifth column provides a magnified view to compare fine detail reconstruction on the fast-moving spray gun.}
    \label{fig:CrossFrameQuality}
    \vspace{-3mm}
\end{figure}

\paragraph{Temporal Partition.}
Our temporal partition strategy performs specialized temporal modeling for highly dynamic 3D Gaussians to reconstruct complex motion details that are averaged out in a single, unified model.
Tab.~\ref{tab:progressive_ablation} and Fig.~\ref{fig:DynamicPartition} clearly demonstrate the quality improvements achieved through temporal partition. 
Notably, even our intermediate partitioning variants ("1.1" and "1.2") achieve significantly superior results compared to the naive temporal slicing baseline, Baseline (seg). This highlights that our dynamic score-based approach is far more effective at improving quality than simply training independent models on fixed temporal segments.

\paragraph{Static Partition.}
Our static partitioning strategy leverages the dynamic score to identify and convert static 3D Gaussians, aiming to improve training efficiency, rendering speed, and storage. As shown in Tab.\ref{tab:progressive_ablation}, static partitioning significantly reduces computational costs compared to using temporal partitioning alone. Furthermore, Tab.\ref{tab:progressive_ablation} and Fig.~\ref{fig:StaticPartition} demonstrate that applying the static partitioning strategy does not cause a decline in reconstruction quality.


\paragraph{Cross-Frame Consistency Loss.}
We introduce the $L_{cross}$ to enforce continuity at boundaries and further enhance overall rendering quality. We quantitatively evaluate temporal consistency using the tOF~\cite{Chu_2020} metric, where lower values indicate better temporal consistency. As shown in Tab.~\ref{tab:progressive_ablation}, we measure tOF over both the entire video (Avg) and at the partition boundaries (Bnd). The data reveals that applying only temporal partitioning, while reducing the average tOF, causes the boundary tOF to rise significantly, which confirms the existence of visual discontinuity at the boundaries. Through the step-by-step integration of $L_{current}$ and $L_{gt}$, the boundary tOF is progressively suppressed, ultimately dropping below the baseline level in our full model and completely mitigating the issue. Our qualitative results further corroborate the dual efficacy of $L_{cross}$. As shown in Fig.~\ref{fig:CrossFrameJitter}, the time-slice visualization intuitively reveals how our method significantly mitigates the severe visual discontinuity at the boundaries. Concurrently, the consecutive frame sequence in Fig.~\ref{fig:CrossFrameQuality} also demonstrates that our full model (+ $L_{gt}$) not only achieves the smoothest transition but also yields additional quality gains.

\paragraph{Analysis of Maximum Partition Level.}
We conduct an ablation study on the maximum partition level to analyze its impact on reconstruction quality and computational cost. As shown in Tab.~\ref{tab:split_level_ablation}, progressively increasing the partition level from 0 to 5 yields a general trend of improved rendering quality. However, we observe diminishing returns in quality gains after level 3, while computational and storage costs continue to increase steadily. Therefore, we select level 3 as the maximum partition level for our main experiments to strike an optimal balance. Furthermore, the results highlight that the cost increase is manageable and not explosive; for instance, the storage at level 4 is merely double that of level 0, demonstrating the efficiency of our scheme.

\begin{table}
\caption{\textbf{Ablation study on the \textit{partition level} parameter.} All experiments are conducted on the \textit{flame\_salmon\_frag3}.}
\vspace{-2mm}
  \label{tab:split_level_ablation}
  \centering
  \resizebox{\columnwidth}{!}{%
    \begin{tabular}{@{}l|cccccc@{}}
      \toprule
      Method & PSNR$\uparrow$ & SSIM$\uparrow$ & LPIPS$\downarrow$ & Storage$\downarrow$ & Training Time$\downarrow$ & FPS$\uparrow$ \\
      \midrule
    0    & 29.93           & 0.923 & 0.61 &  44MB             & 1 hours 13 mins  & 95.21  \\
    1    & 30.08           & 0.927 & 0.56 &  51MB             & 1 hours 24 mins  & 88.13  \\
    2    & 30.21           & 0.932 & 0.54 &  59MB             & 1 hours 37 mins  & 82.81  \\
    3    & 30.30           & 0.934 & 0.52 &  70MB             & 1 hours 56 mins  & 74.58 
    \\
    4    & 30.32           & 0.936 & 0.50 &  88MB          & 2 hours 22 mins  & 64.25 
    \\
    5    & 30.36           & 0.936 & 0.49 &  103MB          & 2 hours 40 mins  & 57.05 
    \\
      \bottomrule
    \end{tabular}
  }
\vspace{-2mm}
\end{table}
\section{Conclusion}

We proposed MAPo, a novel framework for high-fidelity dynamic scene reconstruction. MAPo employs a hierarchical partitioning strategy guided by a dynamic score, enabling specialized modeling for complex motion regions while treating static ones efficiently. To ensure temporal smoothness, we introduce a cross-frame consistency loss to mitigate visual discontinuities at the partition boundaries. Extensive experiments demonstrate that MAPo achieves SOTA rendering quality while maintaining competitive computational efficiency.
{
    \small
    \bibliographystyle{ieeenat_fullname}
    \bibliography{main}
}


\end{document}